%% file: main.tex
\documentclass{article}

\usepackage[final]{corl_2026} % Uncomment for the camera-ready ``final'' version.
\usepackage{amsmath}
\usepackage{amssymb}
\usepackage{wrapfig}
\usepackage{graphicx}
\usepackage{todonotes}

\usepackage{algorithm}
\usepackage{algpseudocode}
\usepackage{multirow}
\usepackage{multicol}
\usepackage{booktabs}
\usepackage{subcaption}
\usepackage{caption}
\usepackage{amsthm}
\usepackage{needspace}
\usepackage{mathtools}

\newtheorem{proposition}{Proposition}

\title{Sensitivity Shaping for Latent Modeling}

\author{
  Hongzhan Yu\textsuperscript{1},
  Chenghao Li\textsuperscript{1},
  Ruipeng Zhang\textsuperscript{1},
  Henrik Christensen\textsuperscript{1},
  Sicun Gao\textsuperscript{1}
  \\
  \textsuperscript{1}University of California San Diego
  \\
  \texttt{\{hoy021, chl235, ruz019, hichrist, sig049\}@ucsd.edu}
}

\begin{document}
\maketitle

%===============================================================================

\begin{abstract}
Generative dynamics models enable planning in challenging systems, 
but safe deployment requires detecting policy-induced out-of-distribution (OOD) transitions.
Existing methods typically treat learned dynamics as fixed and rely on post hoc support surrogates for OOD detection. 
This overlooks a critical failure mode: 
learned dynamics that are insensitive to control changes can map unsupported controls to latent predictions resembling demonstrated transitions, 
suppressing OOD signals despite large prediction errors. 
We introduce support-conditioned control-sensitivity regularization to preserve control-induced variation by promoting local responsiveness in well-supported training regions.
% while limiting unstable extrapolation where empirical support is weak. 
Experiments in vision-based obstacle avoidance, manipulation, and real-robot navigation demonstrate improved OOD detection and safer closed-loop planning.
\end{abstract}


%===============================================================================

\section{Introduction}

Generative models that capture system dynamics through latent representations, such as world models~\cite{ha2018world}, have emerged as promising components for robotics. 
By mapping high-dimensional observations into a latent manifold that preserves transition-relevant structure,
these models support predictive reasoning beyond reactive control.
They have enabled planning and control in challenging systems where analytical dynamics are unavailable,
including vision-based manipulation~\cite{wu2023daydreamer, guo2026flowdreamer} and robot navigation under partial observability in dynamic environments~\cite{shanks2025dreamernav,lai2025world, koh2021pathdreamer}.

A critical challenge is their vulnerability to deployment-time distribution shift~\cite{yu2020mopo,kidambi2020morel}.
Deployed policies often differ from the data-collection policy, changing the state-visitation distribution and querying the model on out-of-distribution (OOD) state-action inputs.
Bounding prediction errors under such OOD queries is difficult. 
Prior work seeks to mitigate this issue by broadening the training distribution, for example through web-scale multimodal data~\cite{radford2021learning,gadre2023datacomp,jia2021scaling,schuhmann2022laion,simeoni2025dinov3} or foundation-model priors~\cite{driess2023palm,ahn2022can,black2026pi0visionlanguageactionflowmodel}. 
However, there is a consensus on the scarcity of robot interaction data relative to other domains~\cite{o2024open, brohan2022rt, zitkovich2023rt, ma2024survey},
and the scale of the training data alone does not guarantee coverage of task-critical counterfactuals or reliable generalization beyond the training manifold.

\begin{figure}
    \centering
    \includegraphics[width=1\linewidth]{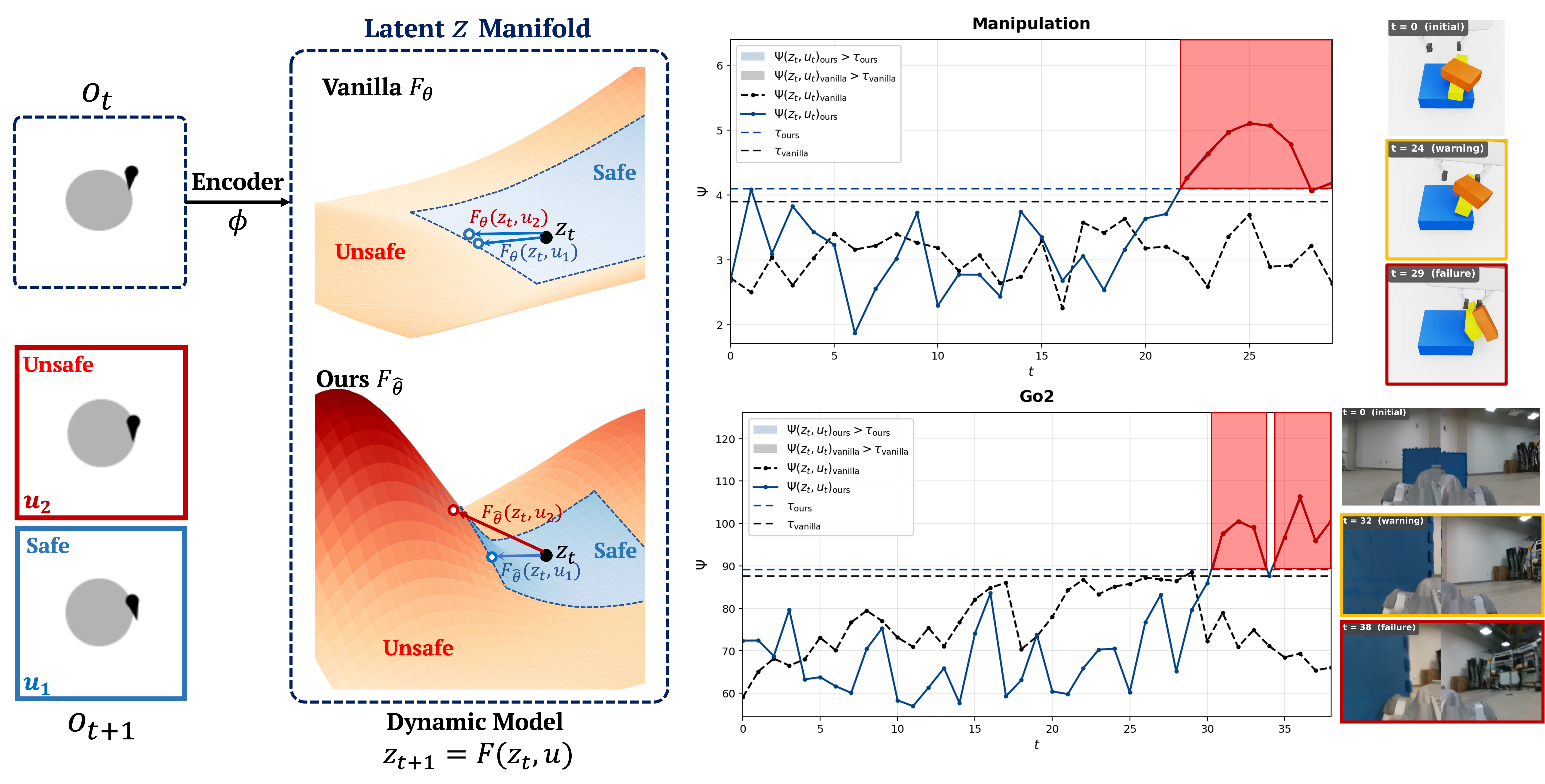}
    \caption{\small 
    Overview of sensitivity shaping for reliable OOD detection in generative dynamics models.
    Control-insensitive models respond weakly to control changes,
    yielding inaccurate yet in-distribution-looking predictions that obscure OOD detection.
    \textbf{(Left)} 
    At a safety-critical state \(o_t\), 
    different controls lead to distinct outcomes, 
    e.g., a safe transition under \(u_1\) and an unsafe one under \(u_2\).
    Yet 
    the vanilla model trained on failure-free data fails to capture these distinct consequences:
    its weak control response flattens the learned dynamics landscape,
    making the predicted next latent states \(z_1\) and \(z_2\) nearly indistinguishable.
    Our sensitivity regularizer promotes nontrivial control responses,
    and improves the detectability of unsupported transitions.
    \textbf{(Right)}
    On the undemonstrated failure rollouts, 
    the vanilla model's scores remain below the calibrated threshold before failure.
    The sensitivity-regularized model flags OOD earlier; 
    red shading marks steps classified as OOD.
    }
    \label{fig:intro}
    \vspace{-0.4cm}
\end{figure}

These limitations motivate uncertainty-aware analysis to detect transitions outside the training support~\cite{DBLP:journals/corr/abs-2110-11334}. 
Existing methods estimate uncertainty through surrogate scores:  
ensembles measure inter-model disagreement~\cite{chua2018deepreinforcementlearninghandful}, 
density models estimate likelihood under a learned data distribution~\cite{lipman2023flowmatchinggenerativemodeling}, 
and nearest-neighbor methods measure distance to observed transitions~\cite{sun2022outofdistributiondetectiondeepnearest}. 
During planning, these scores are incorporated as penalties to discourage unsupported controls~\cite{seo2025uncertainty}.
Such approaches typically assess OOD after learning the dynamics,
using conformal prediction of a surrogate score to bound the marginal false rejection rate for in-distribution inputs under exchangeability~\cite{angelopoulos2021gentle}.
However, reliable OOD-aware control also requires the complementary property: 
 unsupported transitions should be rejected reliably. 
Calibration alone does not provide this property.
It depends on whether the score detects support mismatch along rollout distributions and thus provides an informative proxy for prediction error. 
Importantly, surrogate-measured support mismatch need not align with actual prediction error (Figure \ref{fig:intro}). 

We improve OOD analysis by shaping the local control sensitivity of generative dynamics models during training.
We identify \textit{control insensitivity} as a critical failure mode: 
learned dynamics respond too weakly to control changes and  underrepresent the true responses. 
As a result, 
undemonstrated controls can yield predictions that spuriously resemble the training distribution despite large true predictive errors, 
causing support-based OOD surrogates to miss these transitions.
To address this, we introduce a sensitivity regularizer that targets well-supported regions of the training distribution and promotes nontrivial control-Jacobian Frobenius norms.
This encourages locally responsive dynamics,
while discouraging predictions under different controls from collapsing to overly similar latent transitions,
making unsupported transitions easier to detect.

Overall, we make three contributions. 
First, we analyze control sensitivity in generative dynamics models through a vision-based obstacle-avoidance case study,
demonstrating how control insensitivity degrades OOD detection and formalizing a local failure condition (Section~\ref{section:analysis}). 
Second, we propose a support-conditioned sensitivity regularizer that promotes local responsiveness of the learned dynamics and improves detectability of unsupported transitions
(Section~\ref{section:method}). 
Finally, we show that this method enhances control sensitivity, improves OOD detection, and enables more reliable OOD-aware planning across simulated and real-robot experiments (Section~\ref{section:experiment}).

\textbf{Related Work.}
Generative dynamics models learn latent representations and transition dynamics from high-dimensional observations for prediction, planning, and control~\cite{ha2018world}. 
Recurrent State-Space Models (RSSMs)~\cite{hafner2019learning},
trained with variational objectives~\cite{chung2015recurrent, rezende2014stochastic, kingma2013auto},
form the basis of the Dreamer family~\cite{hafner2019dream,hafner2020mastering,hafner2023mastering, hafner2025training},
which optimizes policies through imagined latent rollouts.
These models have been applied to embodied robotic tasks such as manipulation, navigation, and locomotion~\cite{wu2023daydreamer, shanks2025dreamernav, huang2026pointworld,lai2025world, zhou2024dino}.
However, learned dynamics can become unreliable under deployment-time distribution shift~\cite{hendrycks2016baseline, lakshminarayanan2017simple},
which is especially problematic for safety-critical control~\cite{lambert2020objective, huang2023safedreamer, brunke2022safe}.
Existing out-of-distribution (OOD) analysis methods monitor such failures using distance-based scores~\cite{lee2018simple, wong2022error, liu2024model, fort2021exploring},
density- or energy-based criteria~\cite{nalisnick2018deep, liu2020energy, morningstar2021density, graham2023denoising},
or ensemble uncertainty~\cite{kidambi2020morel, lakshminarayanan2017simple, yu2020mopo},
often to trigger runtime interventions or safety filters~\cite{agia2024unpacking,salehi2021unified,sinha2022system,seo2025uncertainty}.
These post hoc approaches take the learned latent dynamics as given, 
whereas this work shapes them to preserve control-induced variation and improve OOD detection.

% %===============================================================================
\section{Preliminaries}
\label{section:preliminary}

% \subsection{Generative Dynamics Models}
\textbf{Generative Dynamics Models.}
Consider a discrete-time dynamical system.
At each time step $t$,
the agent observes $o_{t} \in \mathcal{O}$,
applies $u_{t} \in \mathcal{U}$,
and transitions to $o_{t+1} = f(o_t, u_t)$ where $f: \mathcal{O} \times \mathcal{U} \to \mathcal{O}$.
Generative dynamics models approximate \(f\) in a lower-dimensional latent space $\mathcal{Z}$ through an encoder posterior and a dynamics prior:
\begin{equation}
\mbox{Encoder:} \hspace{0.2cm} z_{t} \in \mathcal{Z} \sim \phi_{\theta}(\cdot \mid o_{t}), \qquad 
\mbox{Dynamics:} \hspace{0.2cm} \hat{z}_{t+1} \in \mathcal{Z} \sim F_{\theta}(\cdot \mid z_{t}, u_{t}).
\end{equation}
Training promotes latent predictive consistency between the dynamics prior conditioned on $(z_t,u_t)$ and the next-observation encoder posterior:
\(
F_{\theta}(\cdot \mid z_t,u_t) \approx \phi_{\theta}(\cdot \mid o_{t+1}).
\)
Our implementations use Recurrent State-Space Models (RSSMs),
with history conditioning omitted for readability.
For differential analysis, \(F_\theta(z, u)\) denotes a differentiable vector-valued prediction,
and \(\phi_\theta(o)\) an encoded vector.
We write \(d_z\) and \(d_u\) for the dimensions of the latent and continuous control representations.

% \subsection{Out-of-Distribution Analysis}
\textbf{Out-of-Distribution (OOD) Analysis.}
OOD analysis studies whether a learned model is queried outside the support of its training data.
We score latent state-control pairs with surrogate $\Psi: \mathcal{Z} \times \mathcal{U} \rightarrow\mathbb{R}^{+}$,
where larger values indicate lower support.
Common choices include:
\textit{$k$ nearest neighbor} (\({k}\)NN) distances to observed latent transitions~\cite{sun2022outofdistributiondetectiondeepnearest},
\textit{flow matching} scores computed through a learned flow to a Gaussian base~\cite{lipman2023flowmatchinggenerativemodeling},
and \textit{ensemble} disagreement across independently trained dynamics predictors~\cite{chua2018deepreinforcementlearninghandful}.
Appendix~\ref{appendix:ood} specifies all three surrogates. 

Inductive conformal prediction (ICP)~\cite{angelopoulos2021gentle} converts a chosen surrogate into a threshold-based OOD decision rule.
On a held-out in-distribution calibration set,
let \(S_{(1)} \leq \dots \leq S_{(n)}\) denote the ordered scores.
For a prescribed nominal coverage level \(\alpha \in (0, 1)\), 
the finite-sample threshold is
\begin{align}
    k_\alpha = \lceil (n+1)\alpha \rceil, \qquad 
    \tau_\alpha = S_{(k_\alpha)}, \qquad 
    S_{(n+1)} \coloneq +\infty.
\end{align}
A pair \((z_t, u_t)\) is classified as OOD if \(\Psi(z_t, u_t) > \tau_\alpha\).
If calibration and test examples are exchangeable,
the marginal in-distribution acceptance probability is at least \(\alpha\).
This marginal guarantee does not automatically extend to policy-shifted rollouts,
and calibration alone therefore guarantees neither OOD detection nor physical safety.

\begin{figure}[t]
    \centering
    \includegraphics[width=\linewidth]{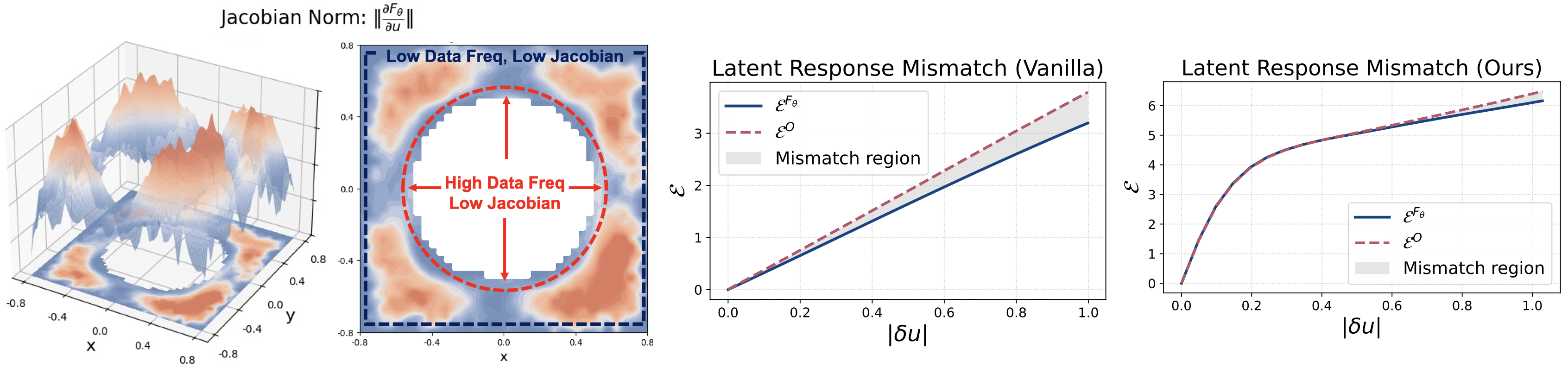}
    \caption{\small
    Sensitivity analysis in the obstacle-avoidance system.
    From left to right:
    \textbf{(a,b)} Control-Jacobian norms for demonstrated latent state-control pairs,
    projected onto the 2D position space and averaged within discretized bins; 
    blue indicates lower sensitivity and red indicates higher sensitivity. 
    Low sensitivity regions coincide with limited control diversity or sparse data coverage.
    \textbf{(c)} The vanilla dynamics underestimate the true encoded response energy even under small control perturbations.
    \textbf{(d)} Sensitivity regularization brings the predicted response energy closer to the true response,
    substantially reducing the gap for small perturbations.
    }
    \label{fig:dubins_jacobian}
    \vspace{-0.2cm}
\end{figure}

% %===============================================================================
\section{Analysis: Control-Conditioned Sensitivity in Latent Dynamics}
\label{section:analysis}

Control insensitivity in the learned dynamics can undermine OOD analysis.
When dynamical predictions change little with the control input, 
they may remain compatible with the training distribution even as their errors grow.
Support-based surrogates can miss these errors because compatibility with training data does not establish transition correctness.
We study this issue in safety-critical systems~\cite{bansal2017hamilton}, 
where failure-free data provide limited control diversity near critical regions.
This weakens the identifiability of control-dependent transitions and induces local control insensitivity in the learned dynamics models.

\subsection{Running Example: Vision-Based Dubins Obstacle Avoidance}

\begin{wrapfigure}{r}{0.55\linewidth}
    \setlength{\intextsep}{4pt}
    \centering
    \vspace{-0.53cm}
    \raisebox{-0.5\height}{%
        \includegraphics[width=\linewidth]{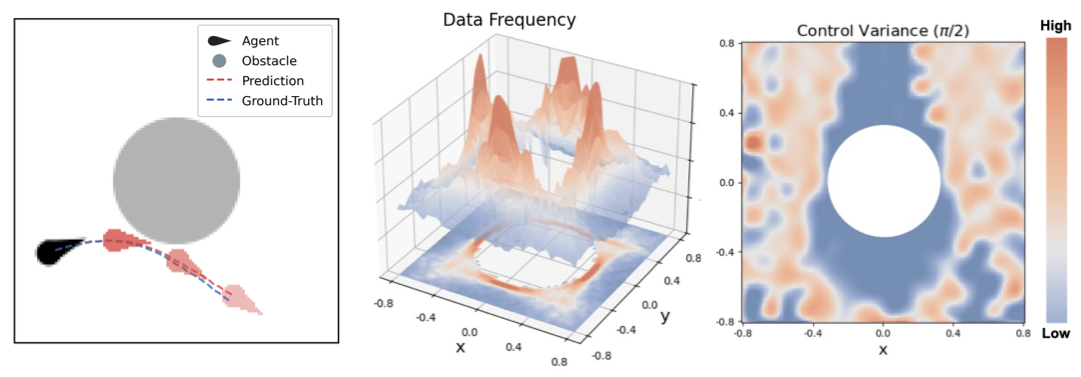}%
    }
    \caption{\small Dubins system and training data.
        \textbf{(Left)} RGB observations with model-predicted future-state overlays tracking the ground-truth.
        \textbf{(Middle)} Spatial training-data density.     
        Coverage is denser near the obstacle and sparser farther away.
        \textbf{(Right)} Control-variance map for heading \(\pi/2\). 
        Low variance near the obstacle indicates limited control diversity under safe constraints.
    }
    \label{fig:dubins_obs_and_variance}
    \vspace{-0.4cm}
\end{wrapfigure}
We illustrate this failure mode in a vision-based obstacle-avoidance case study. 
The agent follows constant-speed Dubins dynamics with a scalar steering-rate input and observes bird's-eye-view images. 
We train generative dynamics models using DreamerV3~\cite{hafner2023mastering} on collision-free trajectories collected with expert and random controls. 
Figure~\ref{fig:dubins_obs_and_variance}-Left shows that the trained model can accurately predict future states for demonstrated state-control pairs.

\textit{\textbf{Training Data Analysis}}.
Figure~\ref{fig:dubins_obs_and_variance}-Middle shows
that samples concentrate near the central obstacle,
while regions farther away are comparatively sparse. 
Despite this substantial state coverage, empirical control variance near the obstacle is low (Figure~\ref{fig:dubins_obs_and_variance}-Right):
excluding failure trajectories restricts near-boundary demonstrations to controls that remain safe.
High state coverage therefore coexists with limited control diversity,
weakening the identifiability of control-dependent dynamics in safety-critical regions.

\subsubsection{Control Sensitivity of the Learned Dynamics}

\textit{\textbf{Measuring Control Sensitivity}}.
We first formalize the sensitivity measure.
For an arbitrary latent state-control pair \((z_t, u_t)\),
we fix \(z_t\) and
view the continuous prediction as a differentiable function in control coordinates: 
\(
    F_{\theta}(z_t, \cdot): \mathbb{R}^{d_u} \to \mathbb{R}^{d_z}.
\)
Its Jacobian
\(
J_uF_\theta(z_t,u_t) 
\in  \mathbb{R}^{d_z \times d_u}
\)
maps a control perturbation direction \(\delta u\) to the first-order change in the predicted next latent state:
\begin{equation}
% \(
\mathrm d F_\theta(z_t, u_t) [\delta u]
=
J_uF_\theta(z_t,u_t) \cdot \delta u
:=
\left.\frac{\mathrm d}{\mathrm d\varepsilon}
F_\theta(z_t, u_t + \varepsilon\,\delta u)
\right|_{\varepsilon=0}.
\end{equation}
We measure control sensitivity by its control-Jacobian Frobenius norm~\cite{novak2018sensitivity}:
\(
\|J_uF_\theta(z_t,u_t)\|_F
=
(
\sum_{i=1}^{d_z}\sum_{j=1}^{d_u}
(J_uF_\theta(z_t,u_t))_{ij}^{2}
)
^{1/2},
\)
aggregating first-order responses across control directions.
This sensitivity measure is a continuous analogue of average sensitivity for Boolean functions~\cite{hahn-2020-theoretical}.

\textit{\textbf{Sensitivity Results}}.
Figure~\ref{fig:dubins_jacobian}-(a,b) visualizes control-Jacobian Frobenius norms over demonstrated state-control pairs.
The results show two low-sensitivity regimes:
sparse outer regions near the position-space boundary,
and near-obstacle regions with substantial state coverage but limited control diversity.
In these regions,
a weak local control response produces similar next-latent predictions for nearby controls.
Yet the true Dubins dynamics (Appendix~\ref{app:dubins}) retain their physical control response near the obstacle:
the steering-rate control input affects the heading with state-independent gain,
while constant speed induces position changes of fixed magnitude.
A fixed control perturbation thus induces a state-independent physical response that does not vanish near the safety boundary.

To quantify this mismatch,
we compare the learned and true control-induced responses.
For an arbitrary latent-valued transition map \(H\) and a feasible perturbation \(\delta u\),
define its control-induced response as 
\(
\Delta_{H}(x, u; \delta u) \triangleq H(x,u+\delta u)
-
H(x, u).
\)
This is the change in \(H\) at fixed \(x\) as control changes from \(u\) to \(u + \delta u\).
Let \(\mathbb{G}: \mathcal{O} \times \mathcal{U} \to \mathcal{O}\) denote the one-step observation transition,
and write \(H_{O}(o, u) \triangleq \phi_{\theta}(\mathbb{G}(o, u))\).
For \((o_t,u_t)\),
with \(z_t = \phi_{\theta}(o_t)\), 
we measure
\begin{align}
\begin{split}
\mathcal{E}^{F_{\theta}}(z_t, u_t; \,\delta u)
\triangleq 
\bigl\|
\Delta_{F_\theta}(z_t, u_t; \,\delta u)\|_2^2, 
\quad
\mathcal{E}^{O}(o_t, u_t; \,\delta u)
\triangleq 
\bigl\|
\Delta_{H_O}(o_t, u_t; \,\delta u)\|_2^2 ,
\label{eq:mismatch_metric}
\end{split}
\end{align}
where \(\mathcal{E}^{F_{\theta}}\) is the model-predicted response and 
\(\mathcal{E}^{O}\) is the true response in the same latent space.  
Figure \ref{fig:dubins_jacobian}-c shows that,
for the near-obstacle, low-sensitivity samples identified in Figure \ref{fig:dubins_obs_and_variance},
the gap between \(\mathcal{E}^{F_{\theta}}\) and \(\mathcal{E}^{O}\) grows with \(\|\delta u\|\),
indicating that learned dynamics increasingly underestimate control-induced variation away from demonstrated controls.
Importantly, the gap is already present for small perturbations, 
revealing a local underestimation of the true response.
These one-step errors can compound during recursive rollouts, undermining the reliability of multi-step predictions.

\subsubsection{Impact on OOD Analysis}
This underestimation can make latent transition geometry misleading for OOD analysis.
If a perturbed control yields a prediction close to demonstrated transitions, 
a support-based score may remain low even when the true transition differs substantially.
The resulting failure reflects the gap between compatibility with the training distribution and predictive accuracy.

\begin{figure}[t]
    \centering
    \includegraphics[width=\linewidth]{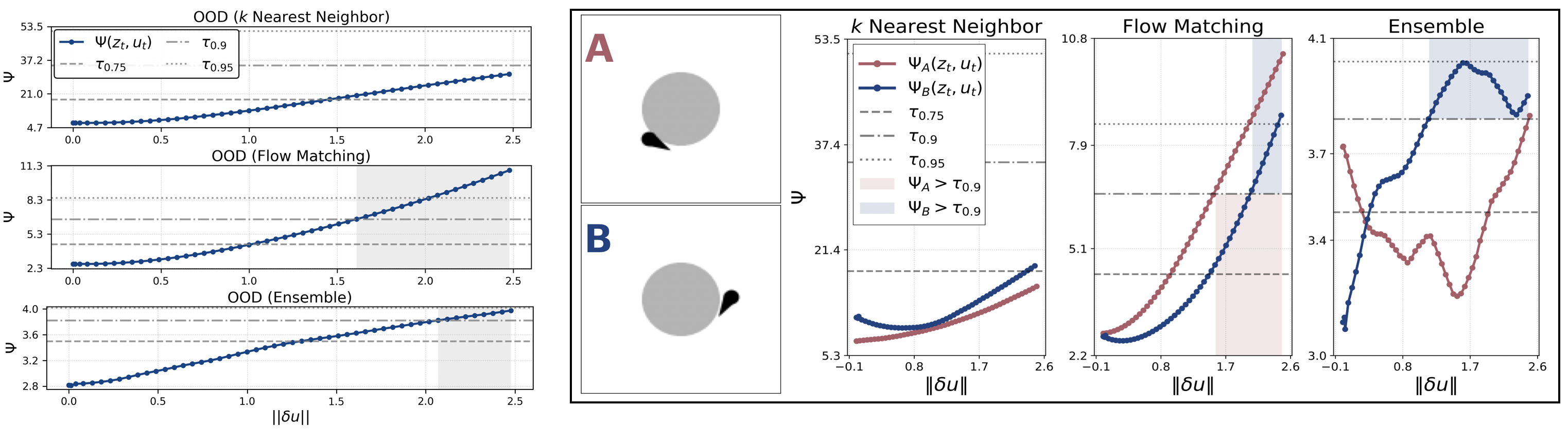}%
    \caption{\small 
    OOD surrogate responses to perturbations.
    \textbf{(Left)} 
    Average surrogate scores over near-obstacle samples as demonstrated controls are perturbed.
    Dashed horizontal lines denote thresholds at coverage levels \(\alpha = 0.75\), \(0.90\), \(0.95\); 
    shaded regions indicate perturbations classified as OOD under \(\tau_{0.90}\).
    \textbf{(Right)}
    Sample-level responses. 
    In both scenarios, the ensemble score decreases over part of the perturbation range.
    }
    \label{fig:dubins_ood1}
    \vspace{-0.4cm}
\end{figure}

\begin{wrapfigure}{r}{0.42\linewidth}
    \setlength{\intextsep}{4pt}
    \centering
    \vspace{-0.48cm}
    \raisebox{-0.5\height}{%
        \includegraphics[width=\linewidth]{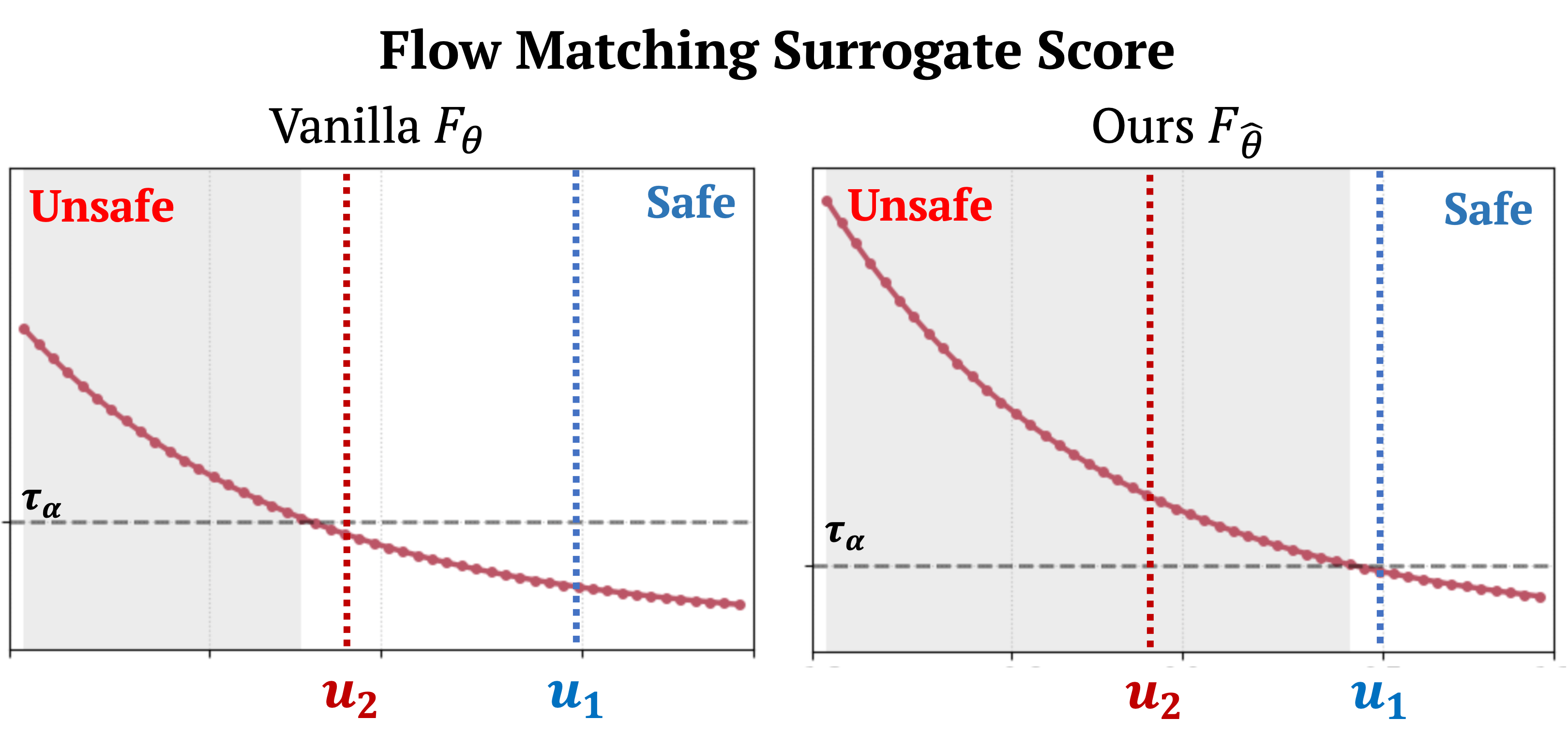}%
    }
    \caption{\small 
    Flow-matching scores for the state \(o_t\) in Figure \ref{fig:intro}.
    Dashed lines indicate calibrated thresholds.
    For the unsafe control \(u_2\),
    vanilla dynamics \textbf{(left)} assigns an in-distribution score due to control-insensitivity,
    whereas the proposed sensitivity-regularized dynamics \textbf{(right)} correctly identifies the transition as OOD.
    }
    \label{fig:dubins_ood2}
    \vspace{-0.2cm}
\end{wrapfigure}
\textit{\textbf{Empirical Results}}.
Over near-obstacle, low-sensitivity samples,
we perturb demonstrated controls by \(\delta u\) and measure OOD scores as \(\|\delta u\|\) increases;
Figure \ref{fig:dubins_ood1} shows that
the scores need not track the growing control deviation.
For \textit{\(k\)NN},
which measures latent distance to a fixed reference set,
control-insensitive predictions can resemble demonstrated transitions and receive small distances.
Its average scores remain below the \(0.90\)-coverage OOD threshold across the full range of \(\|\delta u\|\).
\textit{Ensemble} scores appear more responsive on average but can be non-monotonic at the sample level:
when members share the same control-insensitive bias,
disagreement reflects inter-model variance rather than shared transition error and may even decrease with \(\|\delta u\|\).
\textit{Flow matching} is the most responsive surrogate in this comparison,
but still assigns in-distribution scores to safety-critical perturbations near the training manifold (Figure \ref{fig:dubins_ood2}).
These results show that the post hoc OOD surrogates do not fully compensate for weak control-conditioned variation in the learned dynamics,
motivating targeted regularization of control responsiveness during dynamics training.

\subsection{A Local Failure Condition for OOD Detection}
The analysis suggests a failure mechanism:
a weakly responsive dynamics model can keep its predicted successor inside the OOD detector's acceptance region despite large prediction errors.
We formalize this mechanism through a local failure condition for OOD scores.
Fix an observation \(o\), its history, and \(z=\phi_\theta(o)\). 
Let \(\Psi(z,u)=s_z(F_\theta(z,u))\) be a scalar OOD surrogate score.
For a demonstrated reference control \(u^\ast\in\mathcal U\), 
let
\(d=\|u-u^\ast\|_2\) denote the control deviation,
and define the reference acceptance margin and prediction error as
\begin{align}
    \gamma=\tau_\alpha-\Psi(z,u^\ast),\qquad
    \varepsilon_0=\|H_O(o,u^\ast)-F_\theta(z,u^\ast)\|_2.
\end{align}

\begin{proposition}[Local insensitivity failure] \label{prop:local_insensitivity}
Suppose \(\gamma\in(0,\infty)\) and, for a neighborhood radius \(r>0\), every feasible control with \(d\le r\) satisfies
\begin{equation}
    \|H_O(o,u)-H_O(o,u^\ast)\|_2 \ge md, \qquad
    \|F_\theta(z,u)-F_\theta(z,u^\ast)\|_2 \le L_Fd,
\end{equation}
where \(m>L_F\ge0\) expresses control insensitivity:
the upper sensitivity bound \(L_F\) for the learned response is below the lower expansion bound \(m\) for the true response.
Assume \(s_z\) is \(K\)-Lipschitz on the predicted successors.
For any prediction-error threshold \(E\ge0\), every control satisfying
\begin{equation}
    \frac{E+\varepsilon_0}{m-L_F}
    <d\le
    \min\left\{r,\frac{\gamma}{KL_F}\right\}
    \label{eq:local_failure_interval}
\end{equation}
is accepted, \(\Psi(z,u)\le\tau_\alpha\), despite prediction error
\(\|H_O(o,u)-F_\theta(z,u)\|_2>E\).
The proof is given in Appendix~\ref{app:local_failure}.
\end{proposition}

Every feasible control satisfying (\ref{eq:local_failure_interval}) deviates enough to force prediction error above \(E\), yet remains accepted.
The gap \(m-L_F\) captures the model's insufficient responsiveness relative to the true response and governs growth of the error lower bound,
while the product \(KL_F\) bounds the surrogate's response to control deviations.
A smaller reference error \(\varepsilon_0\) reduces the deviation needed to certify high error, 
whereas the acceptance margin \(\gamma\) specifies the score increase permitted before rejection.
Appendix~\ref{app:local_failure_scores} further extends the analysis to scores with direct control dependence.
Overall, accurate reference predictions do not preclude nearby accepted, high-error controls when learned responsiveness is insufficient relative to the true response.

\section{Method: Support-Conditioned Control Sensitivity Regularization}
\label{section:method}

The preceding study motivates enforcing nontrivial control-Jacobian norms in well-supported regions to counteract the sensitivity mismatch underlying~\eqref{eq:local_failure_interval},
thereby discouraging predictions under unseen controls from collapsing onto demonstrated transitions.

The objective targets OOD detectability,
and it does not guarantee accurate prediction under arbitrary controls.
For a latent state--control pair \((z_t,u_t)\) and valid control perturbations \(\delta u\),
a first-order expansion relates the local response \(\Delta_{F_\theta}(z_t, u_t; \,\delta u) = F_\theta(z, u + \delta u) - F_\theta(z, u)\) to the control Jacobian.
For zero-mean isotropic perturbations with 
\(\sigma_u^2 = \mathbb{E}_{\delta u} \|\delta u\|_2^2 \),
we obtain
\begin{equation}
\mathbb{E}_{\delta u}
% \left[
\|
\Delta_{F_\theta}(z_t, u_t; \,\delta u)
\|_2^2
% \right]
\approx
\mathbb{E}_{\delta u}
% \left[
\left\|
J_uF_\theta(z_t,u_t) \,
\delta u
\right\|_2^2
% \right]
= \frac{{{\sigma}_u^2}}{d_u}  \, \|J_uF_\theta(z_t,u_t)\|_F^2.
\end{equation}
A non-vanishing Frobenius norm indicates aggregate responsiveness but does not  guarantee sensitivity in every direction.
Known perturbation structure, including anisotropy induced by control limits, 
can be incorporated through the second moment, 
yielding the average linearized response
\(
\mathbb E\|J\delta u\|_2^2=\mathrm{tr}(J\,E[\delta u\delta u^\top]\,J^\top).
\)
In this work, we adopt an isotropic reference and regularize aggregate local control sensitivity through the Frobenius norm.

\textit{\textbf{Support-Conditioned Targeting}}.
Enforcing strong responsiveness uniformly across the training manifold can degrade prediction in weakly supported regions:
a global Jacobian constraint may increase local Lipschitz constants and amplify extrapolation errors through sensitive transitions~\cite{zhang2024best}. 
We therefore target high-support regions,
where empirical observations more strongly constrain the latent dynamics. 
We identify these regions by applying the \(k\)-NN surrogate to the demonstrated transitions, 
with distances computed in the predicted-successor latent space.
This training-time assessment ranks demonstrated inputs, 
avoiding control-insensitivity artifacts associated with querying unseen controls.
Its \textit{nonparametric} construction requires no auxiliary density modeling as the representation evolves.

\textit{\textbf{Control Sensitivity Regularization}}.
Since the encoder evolves during training, 
we periodically update the high-support set \(\bar{\mathbb{D}}\).
Each update retains the \(\beta\)-fraction of training transitions with the smallest latent-space \(k\)NN distances.
On this set, we promote local control sensitivity with
\begin{align}
\mathcal{L}_{\mathrm{reg}}
=
\mathbb{E}_{\substack{(o_t,u_t)\sim\bar{\mathbb{D}}\\ z_t\sim \phi_{\theta}(\cdot\mid o_t)}}
\left[
\left(
g
-
\left\|
\frac{\partial F_{\theta}(z_t,u_t)}{\partial u_t}
\right\|_{F}^{2}
\right)_{+}
\right],
\label{eq:reg_obj}
\end{align}
where \((\cdot)_{+}=\max(\cdot,0)\) and \(g>0\) sets a soft lower target on the squared norm.
The hinge penalizes vanishing sensitivity without unbounded growth.
For compute efficiency, we estimate the squared Frobenius norm using Hutchinson’s identity~\cite{hutchinson1989stochastic}.
The training objective is
\(\mathcal{L} = \mathcal{L}_{\mathrm{dyn}} + \lambda_{\mathrm{}{reg}} \mathcal{L}_{\mathrm{reg}},\)
where \(\mathcal{L}_{\mathrm{dyn}}\) denotes the standard generative dynamics training loss and \(\lambda_{\mathrm{}{reg}}\) is the regularization weight.
Appendix~\ref{app:algorithm} gives the training algorithm and the stochastic Jacobian estimator.

\section{Experiments}
\label{section:experiment}
\subsection{Dubins Obstacle Avoidance}
\label{sec:exp_dubins}

We return to the Dubins system (Section~\ref{section:analysis}) to assess control responsiveness and downstream OOD-aware planning.
Appendix~\ref{app:experiments} provides training configurations.

\begin{figure}[t]
    \centering
    \includegraphics[width=\linewidth]{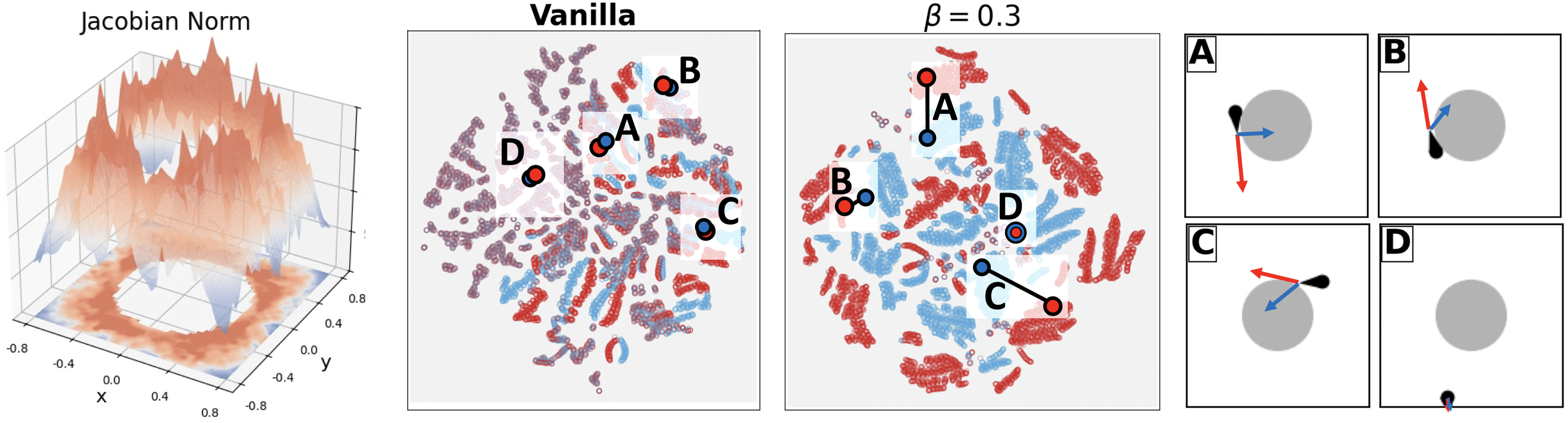}
    \caption{\small 
    Control sensitivity and latent transition geometry after regularization.
    From left to right:
    \textbf{(a)} Control-Jacobian norms of the regularized dynamics,
    showing increased responsiveness in previously control-insensitive regions near the obstacle.
    \textbf{(b,c)} 
    PCA+t-SNE visualizations of latent transitions under demonstrated controls (red) and undemonstrated controls (blue).
    The vanilla model (b) exhibits substantial overlap, 
    indicating underrepresented control-induced variation,
    whereas the regularized model (c) shows improved separation.
    \textbf{(d)} States corresponding to the annotated points in (b,c). 
    }
    \vspace{-0.2cm}
    \label{fig:dubins_exp1}
\end{figure}

\textbf{Sensitivity Analysis.}
Figure~\ref{fig:dubins_exp1}-a shows that,
compared to the unregularized model (Figure~\ref{fig:dubins_jacobian}), 
the sensitivity-regularized model exhibits substantially stronger responsiveness in near-obstacle regions. 
To inspect the induced latent geometry,
Figure~\ref{fig:dubins_exp1}-(b,c) visualizes predicted next-latent states for the top \(20\%\) high-support training samples.
For each sample, we construct an undemonstrated transition by replacing its control with an admissible control farthest from the empirical control set.
Given the pairs \((z_t,u_t)\), 
we evaluate the learned dynamics and project the predicted next-latent states \(z_{t+1}\sim F_{\theta}(\cdot\mid z_t,u_t)\) using PCA followed by t-SNE.
For the unregularized model, 
demonstrated and undemonstrated transitions substantially overlap, 
indicating that control-insensitive dynamics map distinct controls to overly similar latent predictions. 
In contrast, the regularized model yields improved separation despite using a \textit{non-contrastive} Jacobian regularizer.
The remaining overlap typically corresponds to common near-initial or near-terminal states.
Figure~\ref{fig:dubins_jacobian}-d shows that the regularized model also more closely tracks the ground-truth latent response \(\mathcal{E}^{O}\).

\begin{wraptable}{r}{0.4\linewidth}
\vspace{-0.37cm}
\centering
\small
\renewcommand{\arraystretch}{1.12}
\setlength{\tabcolsep}{4pt}
\begin{tabular}{llccc}
\toprule
\textbf{Method} 
& \textbf{$\Psi$}
& $\rho_{\text{succ}} \uparrow$
& $\rho_{\text{filt}}$
& $\rho_{\text{viol}} \downarrow$ \\
\midrule
\multirow{3}{*}{Vanilla}
& $k$NN & 0.600 & 0.340 & 0.273 \\
& FM    & 0.910 & 0.309 & 0.153 \\
& ENS   & 0.380 & 0.459 & 0.327 \\
\midrule
\multirow{3}{*}{Ours}
& $k$NN & 0.800 & 0.323 & 0.166 \\
& FM    & \textbf{0.960} & 0.323 & \textbf{0.104} \\
& ENS   & 0.440 & 0.425 & 0.297 \\
\bottomrule
\end{tabular}
\vspace{2pt}
\caption{\small
Obstacle-avoidance with filtering of a random policy (\(200\) trials).
}
\label{table:dubins_results_a}
\vspace{-0.4cm}
\end{wraptable}

\textbf{OOD-Aware Planning.}
We use calibrated OOD thresholds as one-step safety filters for a random reference controller. 
At each time step, we reject the proposed control \(u_t^{\mathrm{ref}}\) if its induced transition has an OOD score above the calibrated threshold.
Upon rejection, we execute the sampled alternative with the lowest OOD score.
We report: \(\rho_{\mathrm{succ}}\), the fraction of safe trajectories;
\(\rho_{\mathrm{filt}}\), the fraction of rejected reference controls;
and \(\rho_{\mathrm{viol}}\), the fraction of executed transitions classified as OOD.

\underline{\textit{Results}}.
Table~\ref{table:dubins_results_a} compares safety filters built from vanilla and sensitivity-regularized dynamics.
Across all three surrogates, 
regularized dynamics yield higher success rates and lower OOD violation rates than their vanilla counterparts.
Flow matching outperforms \(k\)NN, 
consistent with its greater responsiveness in the preceding analysis.
Ensemble disagreement gives the lowest success rates for both dynamics models,
likely due to spurious inter-member agreement on far-OOD inputs~\cite{fort2021exploring, sun2022outofdistributiondetectiondeepnearest} and observed in Figure~\ref{fig:dubins_ood1}-Right. 
Such agreement is especially problematic for one-step filtering, 
which requires reliable local rankings of candidate controls.
These results support preserving control-induced variation to improve detection of unsupported candidates and OOD-aware filtering.

\underline{\textit{Additional comparison.}}
Inverse-dynamics regularization~\cite{zhang2018decoupling} promotes learned dynamics's control identifiability by training an auxiliary head to recover demonstrated controls from latent transitions.
However, it does not directly constrain forward-model responses to unseen control perturbations.
Our ablations show that this objective does not eliminate insensitivity-induced OOD misclassification.
Support-fraction ablations further show that increasing \(\beta\) improves in-distribution latent prediction but degrades reconstruction fidelity.
Full results are provided in Appendix~\ref{app:ablations}.

\subsection{Block Plucking Manipulation}

\begin{wrapfigure}{r}{0.22\linewidth}
    \setlength{\intextsep}{4pt}
    \centering
    \vspace{-0.48cm}
    \raisebox{-0.5\height}{%
        \includegraphics[width=\linewidth]{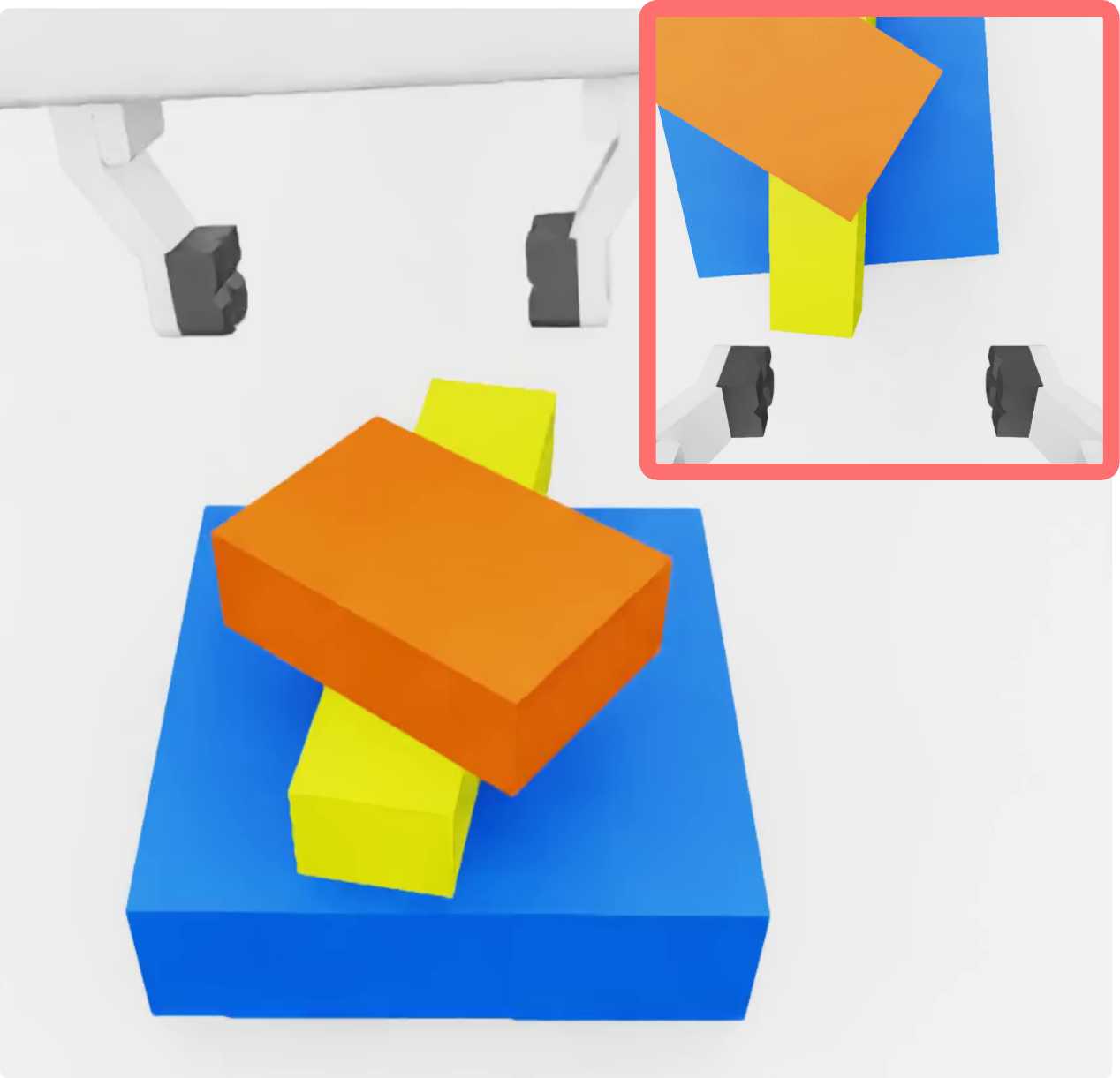}%
    }
    \caption{\small 
    Observations in block-plucking.
    }
    \label{fig:mani}
    \vspace{-0.2cm}
\end{wrapfigure}
\underline{\textit{Setup}}.
We next consider block plucking:
a Franka arm extracts the middle block from a three-block stack without dropping the top block (Figure \ref{fig:mani}). 
Observations comprise RGB images from wrist-mounted and tabletop cameras, 
together with 7-dimensional proprioception. 
The controls are a 6-DoF end-effector pose increment and a discrete gripper command.

We evaluate OOD-aware planning using the ensemble surrogate and a reachability-based safety filter~\cite{seo2025uncertainty}. 
This filter extends the one-step reasoning used in Dubins by learning from imagined latent rollouts to account for long-horizon OOD risk.
Given a surrogate \(\Psi\) and a calibrated threshold \(\tau_\alpha\) at nominal coverage \(\alpha\), 
it learns to avoid trajectories that enter latent states \(z\) from which every admissible control is classified as OOD, i.e.,
\(
\min_{u\in\mathcal{U}} \Psi(z,u) > \tau_\alpha .
\)
A trajectory can remain safe without completing the task,
so we distinguish
the failure rate \(\rho_{\mathrm{fail}}\) from the incompletion rate \(\rho_{\mathrm{inc}}\).
We also report the reconstruction error \(\mathcal{E}_{\text{recon}}\),
the mean summed squared image-reconstruction residual over executed trajectories.
This metric measures how well the model reconstructs executed trajectories.

\underline{\textit{Results}}.
The first comparison holds the reference policy fixed and evaluates filters built from the two dynamics models.
The reference policy has an unfiltered failure rate of  \(\rho_{\mathrm{fail}}\approx 0.25\) (Table~\ref{table:manipulation_result_a}).
The filter trained on sensitivity-regularized dynamics reduces failures relative to the vanilla-dynamics filter with moderate intervention. 
It also produces the lowest reconstruction error, 
indicating more accurate reconstruction of the executed trajectories under the evaluated decoder.

The second comparison evaluates the full offline pipeline:
dynamics and rewards are learned from data,
then both the reward-maximizing reference policy~\cite{hafner2023mastering} and OOD-based safety filter are trained entirely through latent imagination (Table~\ref{table:manipulation_result_b}).
Smaller nominal coverage tightens rejection:
among tested coverage levels,
\(\alpha = 0.80\) minimizes failure and reconstruction error but maximizes incompletion.
At the matched level \(\alpha = 0.90\),
the sensitivity-regularized setting lowers failure with increased intervention and incompletion.
Moreover, the more permissive \(\alpha = 0.95\) still yields lower failure and
intervention than vanilla at \(\alpha = 0.90\),
suggesting that improved control sensitivity preserves safety while reducing the need for overly conservative calibration.

\begin{table*}[t]
\centering
\small
\renewcommand{\arraystretch}{1.12}
\setlength{\tabcolsep}{5pt}

% --- table bodies: bottom-rule aligned ---
\begin{minipage}[b]{0.45\textwidth}
\centering
\raisebox{-5.5pt}{%
\begin{tabular}{lccccc}
\toprule
\textbf{Filter}
& $\alpha$
& $\rho_{\text{fail}} \downarrow$
& $\rho_{\text{inc}}$
& $\rho_{\text{filt}}$
& $\mathcal{E}_{\text{recon}} \downarrow$ \\
\midrule
None
& N/A
& 0.243 & 0.096 & 0.000 & 2335.99 \\
Vanilla
& 0.90 
& 0.196 & 0.047 & 0.242 & 2215.32 \\
Ours
& 0.90 
& \textbf{0.150} & 0.047 & 0.269 & \textbf{2196.11} \\
\bottomrule
\end{tabular}%
}
\end{minipage}
\hfill
\begin{minipage}[b]{0.50\textwidth}
\centering
\begin{tabular}{lccccc}
\toprule
\textbf{Filter}
& $\alpha$
& $\rho_{\text{fail}} \downarrow$
& $\rho_{\text{inc}}$
& $\rho_{\text{filt}}$
& $\mathcal{E}_{\text{recon}} \downarrow$ \\
\midrule
Vanilla
& 0.90
& 0.118 & 0.013 & 0.154 & 2207.55 \\
Ours
& 0.80
& \textbf{0.046} & 0.157 & 0.322 & \textbf{2036.50} \\
Ours
& 0.90
& 0.047 & 0.053 & 0.199 & 2146.66 \\
Ours
& 0.95
& 0.085 & \textbf{0.007} & 0.100 & 2204.37 \\
\bottomrule
\end{tabular}
\end{minipage}

\vspace{4pt}

% --- captions: separate from table-body alignment ---
\begin{minipage}[t]{0.45\textwidth}
\centering
\captionof{table}{\small
Block-plucking with filtering of a suboptimal policy (\(1000\) trials).
}
\label{table:manipulation_result_a}
\end{minipage}
\hfill
\begin{minipage}[t]{0.5\textwidth}
\centering
\captionof{table}{\small
Block-plucking with filtering of the policy trained from the same dynamics model (\(1000\) trials).
}
\label{table:manipulation_result_b}
\end{minipage}
\vspace{-0.6cm}
\end{table*}

\subsection{Real-Robot Static-Obstacle Avoidance}

\begin{wrapfigure}{r}{0.38\linewidth}
    \setlength{\intextsep}{4pt}
    \centering
    \vspace{-0.48cm}

    \includegraphics[width=\linewidth]{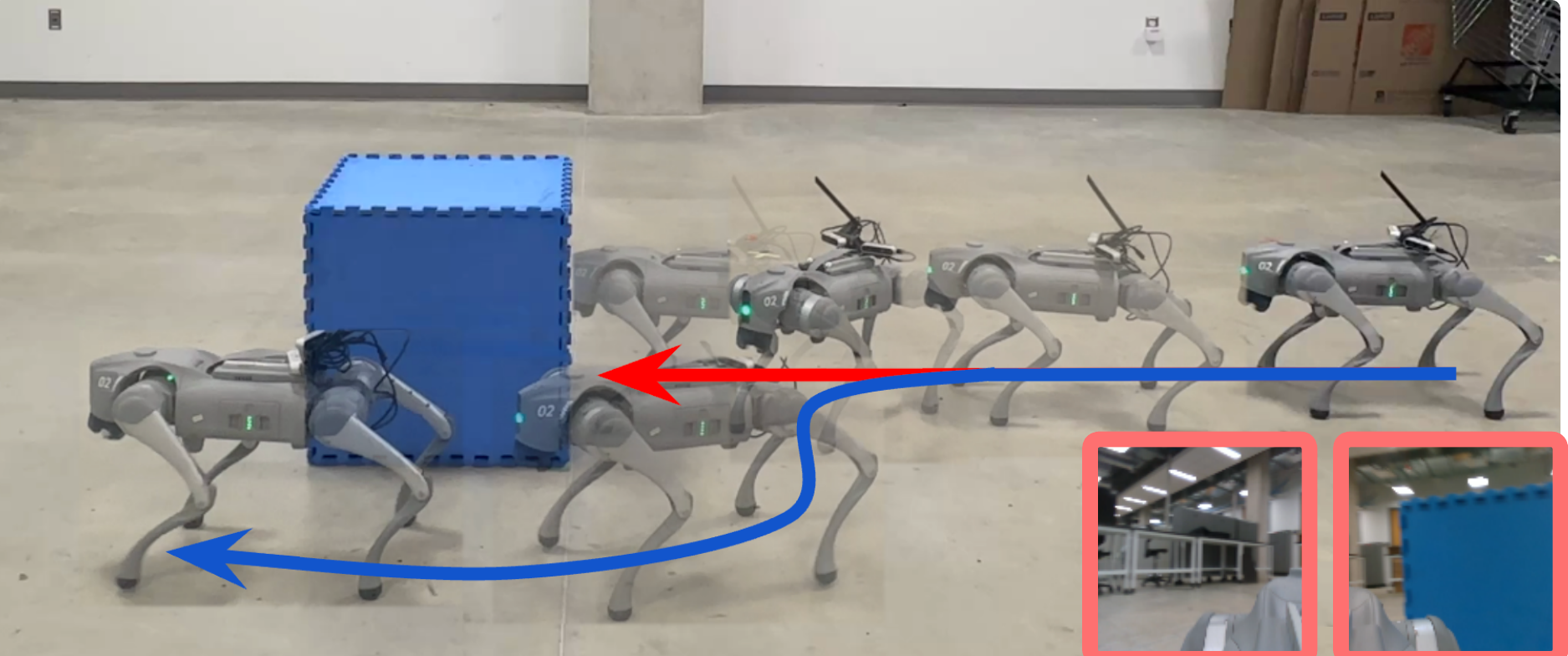}
    \captionof{figure}{\small
    Real-robot obstacle avoidance system.
    The unfiltered trajectory (red) results in failure,
    whereas the flow-matching filter (blue) avoids it.
    Bottom-right insets show body-camera observations.
    }
    \label{fig:go2_setup}

    \vspace{0.2cm}

    \small
    \renewcommand{\arraystretch}{1.12}
    \setlength{\tabcolsep}{5pt}
    \resizebox{\linewidth}{!}{%
    \begin{tabular}{llccc}
    \toprule
    \textbf{Method}
    & \textbf{$\Psi$}
    & $\rho_{\text{succ}} \uparrow$
    & $\rho_{\text{filt}}$
    & $\rho_{\text{viol}} \downarrow$ \\
    \midrule
    \multirow{3}{*}{Vanilla}
    & $k$NN & 0.240 & 0.047 & 0.027 \\
    & FM    & 0.900 & 0.343 & 0.155 \\
    & ENS   & 0.150 & 0.543 & 0.507 \\
    \midrule
    \multirow{3}{*}{Ours}
    & $k$NN & 0.380 & 0.040 & \textbf{0.026} \\
    & FM    & \textbf{1.000} & 0.326 & 0.154 \\
    & ENS   & 0.750 & 0.592 & 0.557 \\
    \bottomrule
    \end{tabular}
    }
    \captionof{table}{\small
    Real-robot with filtering of a random policy (\(50\) trials).
    }
    \label{table:go2_results}
    \vspace{-0.4cm}
\end{wrapfigure}
\underline{\textit{Setup}}.
Lastly, we study a real-robot obstacle-avoidance system
using images from two body-mounted cameras (Figure~\ref{fig:go2_setup}). 
We collect \(60\) minutes of collision-free teleoperation
using three discrete commands: turn left, move straight, and turn right,
all executed at a fixed forward speed of \(0.5\,\mathrm{m/s}\).
Although commands are discrete, learned dynamics can still be insensitive to command changes as this is a data-induced limitation of learned dynamics.
We treat the one-hot command vector as a continuous relaxation: the dynamics Jacobian with respect to its coordinates serves as the surrogate for command-conditioned responsiveness.

\underline{\textit{Results}}.
Table~\ref{table:go2_results} reports filtering results for a random reference policy,
using one-step filters for \(k\)NN and flow matching surrogates and a reachability-based filter for ensemble.
Limited \(k\)NN success is consistent with the difficulty of using latent distances for real-world visual variation that cannot be exhaustively covered.
Flow matching achieves the highest success within the data budget and improves with regularization.
This suggests an advantage of the parametric flow-based surrogate over direct latent-space geometric measures under limited coverage.
The ensemble filter intervenes frequently, 
as scene changes can induce observation-level disagreement that compounds over long-horizon imagined rollouts.
Nevertheless, the large \(\rho_{\mathrm{succ}}\) improvement suggests that reachability filtering benefits from better control-conditioned action discrimination.
We interpret \(\rho_{\mathrm{viol}}\) cautiously: 
unmodeled observation-level shifts can affect the rate but lie outside our focus on control-induced OOD.
Violations are also defined relative to the calibrated support surrogate, 
not ground-truth support or physical safety.
Overall, we show that regularization improves hardware success rates across all evaluated surrogates and identify robustness to observation-level shifts as a key future direction for broader real-world deployment.

\section{Conclusion}
\label{sec:conclusion}
We identified a failure mode in which control-insensitive dynamics models map unsupported controls to predictions resembling demonstrated transitions, 
suppressing OOD signals despite large prediction errors.
We formalized a local failure condition for successor-based OOD scores and introduced support-conditioned control-sensitivity regularization to mitigate this failure.

Across several vision-based tasks, our method increases control responsiveness, improves OOD detection, and enables safer closed-loop planning.
These results show that OOD detection depends not only on post hoc support estimation but also on whether learned dynamics preserve the control-induced variation needed to distinguish unsupported queries.

\textbf{Limitations.}
Our method targets control-induced detection failures and does not guarantee accurate extrapolation under arbitrary distribution shift.
Periodic support updates and Jacobian estimation add training overhead (Appendix~\ref{app:training_cost}), 
with support updates becoming costly for large datasets.
The method's effectiveness depends on where and how sensitivity is shaped:
latent \(k\)-NN support estimates inherit the geometry of the learned representation, 
while the Frobenius-norm objective promotes aggregate responsiveness without ensuring faithful responses in every control direction.
Our formal analysis establishes a local failure condition for successor-based scores, not a general characterization of OOD detection.
Real-robot experiments further identify observation-level shifts as a distinct challenge.
Scalable support estimation, direction-aware regularization, and explicit treatment of observation-level shifts remain important directions for broader real-world deployment.

\bibliography{references}  % .bib

%===============================================================================

\clearpage
% The acknowledgments are automatically included only in the final and preprint versions of the paper.
% \acknowledgments{}
\include{appendix}

\end{document}

%% file: appendix.tex
\appendix

\section{OOD Surrogates and Calibration}
\label{appendix:ood}

This section specifies the three OOD surrogates and the conditions under which their calibrated thresholds provide marginal in-distribution coverage.
All three assign a real-valued nonconformity score \(\Psi(z,u)\); a query is accepted when \(\Psi(z,u)\le\tau_\alpha\).

\subsection{Representations and Data Roles}
\label{app:ood_data}

Let \(\mathbb D=\{(o_t^{(i)},u_t^{(i)},o_{t+1}^{(i)})\}_{i=1}^N\) be the dynamics-training data, with the observation history supplied to the recurrent model.
Once dynamics training is complete, the encoder and dynamics parameters are frozen for OOD-surrogate fitting and evaluation.
For parametric surrogate fitting, write
\begin{equation}
  \mathbb Z_{\mathrm{fit}}=\{(z_t^{(i)},u_t^{(i)},\bar z_{t+1}^{(i)})\}_{i=1}^{N}
\end{equation}
for latent transitions computed from \(\mathbb D\), 
where \(z_t^{(i)}\) is inferred from the observation history and \(\bar z_{t+1}^{(i)}\) is a successor representation produced by the learned dynamics.
Both flow matching and the ensemble are fitted on these training transitions.
When the successor is sampled, its auxiliary randomness is part of the score construction and is thus treated consistently during calibration and testing.

Training-time support selection and deployment-time OOD analysis use data for different purposes.
The training-time \(k\)NN gate ranks demonstrated transitions against the training data and selects where to apply regularization.
Deployment \(k\)NN uses a fixed reference set of predicted successors from the training data as well.
A separate calibration dataset is used only to select the thresholds \(\tau_\alpha\) after the dynamics, parametric surrogates, and reference set have been fixed.
Thus training data determine the score, while calibration data determine its rejection threshold.
The exchangeability conditions for the marginal coverage statement are given in Section~\ref{app:calibration}.

\subsection{Nearest-Neighbor Surrogate}
\label{app:knn}

The \(k\)NN surrogate~\cite{sun2022outofdistributiondetectiondeepnearest} measures the distance from a predicted successor \(\bar z\) to a fixed reference set \(\mathcal R=\{z_i\}_{i=1}^{N_{\mathrm{ref}}}\).
Let \(\mathcal N_k(\bar z;\mathcal R)\) index its \(k\) nearest reference successors, with \(1\le k\le N_{\mathrm{ref}}\).
Using Euclidean distances, the score is
\begin{equation}
  \Psi^{\mathrm{kNN}}(z,u)
  =\frac1k\sum_{i\in\mathcal N_k(\bar z;\mathcal R)}\|\bar z-z_i\|_2.
  \label{eq:app_knn}
\end{equation}
Larger scores indicate that the prediction is farther from the reference successors.
% The reference set remains fixed during evaluation.
This construction is a geometric surrogate for support: closeness of predicted successors does not establish that the queried control was demonstrated at the current state.
For a fixed reference set, this score is \(1\)-Lipschitz in \(\bar z\) and
this bound is in the coordinates of the scored successor.

\subsection{Flow-Matching Surrogate}
\label{app:fm}

The flow-matching surrogate~\cite{lipman2023flowmatchinggenerativemodeling} learns a conditional transport from predicted-successor representations to a Gaussian base distribution.
We use \(\vartheta\) for surrogate parameters to distinguish them from the reference acceptance margin \(\gamma\) in the main proposition.
The conditional vector field is
\begin{equation}
  v_\vartheta:\mathbb R^{d_z}\times[0,1]\times(\mathcal Z\times\mathcal U)
  \longrightarrow\mathbb R^{d_z}.
\end{equation}
For \((z,u,\bar z)\sim\mathbb Z_{\mathrm{fit}}\), draw \(s\sim\mathrm{Unif}([0,1])\) and \(\epsilon\sim\mathcal N(0,I_{d_z})\), and interpolate as
\(\tilde z_s=(1-s)\bar z+s\epsilon\).
The target velocity points from the successor toward the Gaussian sample, giving the training objective
\begin{equation}
  \min_\vartheta\;
  \mathbb E_{(z,u,\bar z),s,\epsilon}
  \left[\left\|v_\vartheta(\tilde z_s,s\mid z,u)-(\epsilon-\bar z)\right\|_2^2\right].
  \label{eq:app_fm_training}
\end{equation}
To evaluate a query, initialize \(y^{(0)}=\bar z\) and use \(L=10\) forward Euler steps with \(\Delta s=1/L\):
\begin{equation}
  y^{(\ell+1)}=y^{(\ell)}+
  \Delta s\,v_\vartheta(y^{(\ell)},\ell\Delta s\mid z,u),
  \qquad \ell=0,\ldots,L-1.
\end{equation}
The surrogate is the Gaussian base energy, including its normalization constant:
\begin{equation}
  \Psi^{\mathrm{FM}}(z,u)
  =-\log\mathcal N(y^{(L)};0,I_{d_z})
  =\frac12\|y^{(L)}\|_2^2+\frac{d_z}{2}\log(2\pi).
  \label{eq:app_fm_score}
\end{equation}
Omitting the constant and calibrating the resulting score equivalently shifts the threshold without changing the decision rule.

This base-energy score is a support heuristic, not the negative log-likelihood of the predicted successor.
For a continuous flow with a well-defined density, the exact change-of-variables expression would be~\cite{chen2018neural}
\begin{equation}
  -\log p_\vartheta(\bar z\mid z,u)
  =-\log p_{\mathrm{base}}(y_1)
  -\int_0^1\nabla_y\!\cdot v_\vartheta(y_s,s\mid z,u)\,\mathrm ds,
  \label{eq:fm_exact_density}
\end{equation}
where \(y_0=\bar z\) and \(\mathrm dy_s/\mathrm ds=v_\vartheta(y_s,s\mid z,u)\).
Our score omits the divergence term and uses finite-step integration, so it need not preserve successor-density rankings.
The base-energy score also remains defined for vector representations of discrete latents, for which an ordinary continuous density interpretation requires additional assumptions.
Finally, the flow is conditioned directly on \(u\) and its score need not vary only through \(F_\theta(z,u)\).

\subsection{Ensemble Surrogate}
\label{app:ensemble}

The ensemble surrogate~\cite{chua2018deepreinforcementlearninghandful} measures disagreement among \(M=10\) independently trained predictors that imitate the frozen latent dynamics.
Member \(m\) predicts a Gaussian distribution
\begin{equation}
  q_{\vartheta_m}(\cdot\mid z,u)
  =\mathcal N\bigl(\mu_{\vartheta_m}(z,u),\Sigma_{\vartheta_m}(z,u)\bigr),
  \qquad \Sigma_{\vartheta_m}(z,u)\succ0,
\end{equation}
Each Gaussian head predicts a mean and element-wise log standard deviation \(b_m\), with \(b_m\) clipped to \([-20,1]\) and \(\Sigma_{\vartheta_m}=\operatorname{diag}(\exp(2b_m))\).
The predictor head has four linear layers,
% four hidden layers of widths \(h,2h,3h,h\), 
each with layer normalization and ReLU, followed by a linear output layer of width twice the prediction dimension.
% \(h\) equals the input dimension.
Each member is fitted to the model-predicted successors using
\begin{equation}
  \min_{\vartheta_m}\;
  \mathbb E_{(z,u,\bar z)\sim\mathbb Z_{\mathrm{fit}}}
  [-\log q_{\vartheta_m}(\bar z\mid z,u)].
\end{equation}
The equal-weight predictive mixture is
\(\bar q(\cdot\mid z,u)=M^{-1}\sum_{m=1}^M q_{\vartheta_m}(\cdot\mid z,u)\).
We use the order-two R\'enyi entropy and analytic Gaussian-overlap evaluation.
Suppressing the conditioning variables, define
\begin{equation}
  \mathcal H_2(q)=-\log\int q(y)^2\,\mathrm dy,
  \qquad C_{ij}=\int q_i(y)q_j(y)\,\mathrm dy
  =\mathcal N(\mu_i;\mu_j,\Sigma_i+\Sigma_j).
\end{equation}
The Jensen--R\'enyi entropy gap is therefore
\begin{align}
  D_2(z,u)
  &=\mathcal H_2(\bar q)-\frac1M\sum_{m=1}^M\mathcal H_2(q_m)\nonumber\\
  &=-\log\!\left(\frac1{M^2}\sum_{i,j=1}^M C_{ij}\right)
  +\frac1M\sum_{i=1}^M\log C_{ii}.
  \label{eq:app_ensemble}
\end{align}
We use its magnitude \(|D_2|\) as the disagreement score, with no Monte Carlo entropy estimate required.
For positive magnitudes, a logarithmic score representation preserves the calibrated acceptance decision when the threshold is transformed consistently.
Larger disagreement scores are treated as a heuristic indication of lower support.

\subsection{Coverage-Indexed Calibration}
\label{app:calibration}

We use \(\alpha\) as the nominal in-distribution \emph{coverage}.
For a score fixed before threshold calibration, let \(S_i=\Psi(z_i,u_i)\), \(i=1,\ldots,n\), be the calibration scores.
Order them as \(S_{(1)}\le\cdots\le S_{(n)}\), 
append \(S_{(n+1)}=+\infty\), 
and set
\begin{equation}
  k_\alpha=\lceil(n+1)\alpha\rceil,
  \qquad \tau_\alpha=S_{(k_\alpha)},
  \qquad \alpha\in(0,1).
  \label{eq:app_conformal}
\end{equation}
For a new in-distribution query with score \(S_{n+1}\), exchangeability of the calibration and test scores, conditional on the fitted scorer, gives
\begin{equation}
  \Pr\{S_{n+1}\le\tau_\alpha\}\ge\alpha.
\end{equation}
To see this, first break ties randomly: exchangeability makes the test score's rank uniform among \(n+1\) scores, and the threshold accepts at least the first \(k_\alpha\) ranks.
Accepting equality without random tie-breaking can only increase coverage.
Thus the false-rejection probability is at most \(1-\alpha\).

The fitted scorer includes every data-dependent component used to evaluate a query: the encoder, dynamics, flow or ensemble parameters, and the \(k\)NN reference set.
Ordinary split conformal requires these components to be fixed independently of the examples reserved for threshold calibration~\cite{angelopoulos2021gentle}.
In our implementation, training data supply both the nearest-neighbor references and the fitting examples for flow matching and ensemble predictors; separate calibration data are used only for threshold selection.
Conditional on the training data and fitted score, the stated marginal guarantee applies when calibration and test scores are exchangeable.

For fixed scores, increasing \(\alpha\) weakly increases \(\tau_\alpha\) and makes rejection more permissive.
The coverage statement is marginal over exchangeable in-distribution queries; it is not a conditional guarantee at every latent state and does not automatically apply to adaptively selected controls or policy-shifted trajectories.
It provides neither an OOD-detection rate nor a physical safety guarantee.

\section{Training Algorithm and Jacobian Estimation}
\label{app:algorithm}
\label{appendix:alg}

\subsection{Support-Conditioned Training}

Algorithm~\ref{alg:pseudo} augments the standard dynamics loss with the regularizer in Eq.~\eqref{eq:reg_obj}.
We first train the model without regularization for \(N_{\mathrm{start}}\) steps.
At the first regularized step, and every \(N_{\mathrm{update}}\) steps thereafter, we recompute predicted successors for the demonstrated training transitions and rank their training-set \(k\)NN distances.
The high-support set \(\bar{\mathbb D}\) retains the \(\beta\)-fraction with the smallest scores.
Support-set membership is held fixed between refreshes, while the model continues to learn.
The gate ranks observed training transitions; it does not require sampled unseen controls or counterfactual successor labels.
Its nonparametric construction avoids fitting an auxiliary density model at each refresh, although it still inherits the geometry of the evolving latent representation.

\begin{algorithm}[htbp]
\caption{Support-conditioned control-sensitivity training}
\label{alg:pseudo}
\begin{algorithmic}[1]
\Require Demonstration dataset \(\mathbb D\); support fraction \(\beta\in(0,1]\); squared-norm target \(g>0\); regularization weight \(\lambda_{\mathrm{reg}}\ge0\); neighbor count \(k\); probe count \(R\); warm-up steps \(N_{\mathrm{start}}\); refresh interval \(N_{\mathrm{update}}\); total steps \(N_{\mathrm{total}}\)
\State Initialize encoder and latent dynamics parameters \(\theta\)
\For{\(n=1,\ldots,N_{\mathrm{total}}\)}
  \State Sample a training minibatch \(\mathcal B\) from \(\mathbb D\)
  \State Compute the base generative dynamics loss \(\mathcal L_{\mathrm{dyn}}(\mathcal B;\theta)\)
  \State \(\mathcal L\gets\mathcal L_{\mathrm{dyn}}\)
  \If{\(n>N_{\mathrm{start}}\)}
    \If{\((n-N_{\mathrm{start}}-1)\bmod N_{\mathrm{update}}=0\)}
      \State Recompute demonstrated predicted successors with the current model
      \State Rank training transitions by their training-set \(k\)NN scores
      \State \(\bar{\mathbb D}\gets\) the \(\beta\)-fraction with the smallest scores \label{algline:update_support}
    \EndIf
    \State Sample a regularization minibatch \(\mathcal B_{\mathrm{reg}}\) from \(\bar{\mathbb D}\)
    \For{each \((o_t,u_t,o_{t+1})\in\mathcal B_{\mathrm{reg}}\)}
      \State Infer \(z_t\) from its observation history
      \State Draw independent probes \(v_1,\ldots,v_R\sim\mathcal N(0,I_{d_u})\)
      \State Compute \(w_j=J_uF_\theta(z_t,u_t)v_j\), holding \(z_t\) and its history fixed
      \State \(\widehat S_t\gets R^{-1}\sum_{j=1}^R\|w_j\|_2^2\)
    \EndFor
    \State \(\widehat{\mathcal L}_{\mathrm{reg}}\gets |\mathcal B_{\mathrm{reg}}|^{-1}\sum_{t\in\mathcal B_{\mathrm{reg}}}(g-\widehat S_t)_+\)
    \State \(\mathcal L\gets\mathcal L_{\mathrm{dyn}}+\lambda_{\mathrm{reg}}\widehat{\mathcal L}_{\mathrm{reg}}\) \label{algline:apply}
  \EndIf
  \State Update \(\theta\) using \(\nabla_\theta\mathcal L\)
\EndFor
\end{algorithmic}
\end{algorithm}

\subsection{Continuous Control Jacobian}
\label{app:jacobian_estimator}

The differentiated map \(F_\theta(z,u)\) is the vector of next-state prior logits, and the Jacobian is
\begin{equation}
  J_uF_\theta(z,u)=\frac{\partial F_\theta(z,u)}{\partial u}
  \in\mathbb R^{d_z\times d_u}.
\end{equation}
This is a partial derivative with respect to the current control input: the current latent state, observation, and recurrent history are held fixed when forming it.
It does not differentiate a newly observed successor through the physical system,
but measures sensitivity of the model's input representation.

\subsection{Squared-Norm Estimator and Hinge Objective}

Writing \(J=J_uF_\theta(z,u)\), Hutchinson's identity~\cite{hutchinson1989stochastic} gives
\begin{align}
  \mathbb E_{v\sim\mathcal N(0,I_{d_u})}\|Jv\|_2^2
  &=\mathbb E_v\,\mathrm{tr}(Jvv^\top J^\top)\\
  &=\mathrm{tr}(JJ^\top)=\|J\|_F^2.
\end{align}
With \(R=5\) independent Gaussian probes per regularization query, we use
\begin{equation}
  \widehat S=\frac1R\sum_{j=1}^R\|Jv_j\|_2^2,
  \qquad \mathbb E[\widehat S\mid J]=\|J\|_F^2.
  \label{eq:app_hutchinson}
\end{equation}
Each term requires a Jacobian--vector product, avoiding explicit construction of the full \(d_z\times d_u\) Jacobian.
Optimizing the regularizer requires retaining the differentiation graph through these products when computing parameter gradients.
The probes estimate aggregate Jacobian energy; they are not controls applied to the system and need not satisfy the physical control constraints.

The implemented minibatch penalty uses \((g-\widehat S)_+\).
Thus \(g\) is a soft target for the \emph{squared} Frobenius norm, rather than a hard lower bound on the learned Jacobian or a target for its unsquared norm.
Above the target, the exact hinge \((g-\|J\|_F^2)_+\) provides no further incentive to increase sensitivity; it does not impose an upper bound either.
Although Eq.~\eqref{eq:app_hutchinson} is unbiased, applying the hinge to the estimate is not:
\begin{equation}
  \mathbb E[(g-\widehat S)_+\mid J]
  \ge (g-\|J\|_F^2)_+,
\end{equation}
by convexity.
Finally, the Frobenius norm aggregates all singular directions and cannot ensure a nonzero response in every control direction.

\subsection{Perturbation Geometry and Coordinate Scale}
\label{app:perturbation_geometry}

For a control perturbation distribution with finite second moment
\(C_u=\mathbb E[\delta u\delta u^\top]\), the linearized response satisfies
\begin{equation}
\mathbb E\|J\delta u\|_2^2=\mathrm{tr}(JC_uJ^\top).
\end{equation}
This identity does not require zero mean; \(C_u\) is the second moment, not necessarily the covariance.
For zero-mean isotropic perturbations with \(\mathbb E\|\delta u\|_2^2=\sigma_u^2\), it reduces to
\(\sigma_u^2\|J\|_F^2/d_u\), giving the first-order approximation in the main paper.
The approximation to the nonlinear response is local and need not hold for large control changes.
Actual feasible perturbations can have unequal directional moments, particularly near a control limit; their expected linearized response is described by \(C_u\).
Our regularizer instead uses isotropic computational probes to measure aggregate responsiveness without choosing task-specific perturbation directions.

\section{Local Insensitivity and Accepted Prediction Error}
\label{app:local_failure}
\label{appendix:theory}

We prove Proposition~\ref{prop:local_insensitivity} and describe its scope.
Throughout this section, the observation $o$, its history, the latent state
$z=\phi_\theta(o)$, and the fitted models are fixed. The maps $H_O(o,u)$ and
$F_\theta(z,u)$ denote the posterior and prior successor logits, respectively,
in matching coordinates. The proposition concerns
these deterministic summaries. All vector norms are Euclidean, and $\|\cdot\|_{\mathrm{op}}$ denotes
the induced operator norm. Acceptance means $\Psi(z,u)\le\tau_\alpha$.

\subsection{Proof and Interpretation}
\label{app:local_failure_proof}

\begin{proof}[Proof of Proposition~\ref{prop:local_insensitivity}]
Let $u\in\mathcal U$ satisfy $d=\|u-u^\ast\|_2\le r$. The score's
$K$-Lipschitz property and the learned-response bound imply
\begin{align}
\Psi(z,u)
&\le \Psi(z,u^\ast)
   +K\|F_\theta(z,u)-F_\theta(z,u^\ast)\|_2\nonumber\\
&\le \tau_\alpha-\gamma+KL_Fd.
\label{eq:app_local_score_bound}
\end{align}
The reverse triangle inequality gives
\begin{align}
\|H_O(o,u)-F_\theta(z,u)\|_2
&\ge \|H_O(o,u)-H_O(o,u^\ast)\|_2
     -\|F_\theta(z,u)-F_\theta(z,u^\ast)\|_2\nonumber\\
&\hspace{1em}-\|H_O(o,u^\ast)-F_\theta(z,u^\ast)\|_2\nonumber\\
&\ge (m-L_F)d-\varepsilon_0.
\label{eq:app_local_error_bound}
\end{align}
Thus $d\le\gamma/(KL_F)$ ensures acceptance, whereas
$d>(E+\varepsilon_0)/(m-L_F)$ ensures prediction error greater than $E$.
When $KL_F=0$, Equation~\eqref{eq:app_local_score_bound} guarantees acceptance
throughout the neighborhood; we therefore interpret $\gamma/0=+\infty$.
Combining these inequalities with $d\le r$ proves the claim.
\end{proof}

\paragraph{Existence of a failure.}
Define
\begin{equation}
R=\min\left\{r,\frac{\gamma}{KL_F}\right\},\qquad
\mathcal A_E=
\left\{u\in\mathcal U:
\frac{E+\varepsilon_0}{m-L_F}<\|u-u^\ast\|_2\le R\right\}.
\label{eq:app_failure_set}
\end{equation}
Every control in $\mathcal A_E$ is accepted despite error exceeding $E$.
An actual failure is certified when $\mathcal A_E\ne\varnothing$.
For a general feasible-control set, the strict inequality
$(E+\varepsilon_0)/(m-L_F)<R$ is necessary but does not alone ensure that
a feasible control occurs at such a distance. It is sufficient if a feasible
ray from $u^\ast$ extends to distance $R$. The strict lower endpoint matters:
at equality, the error bound only establishes error at least $E$.
An empty certified set does not imply reliable detection.

\paragraph{Roles of the four quantities.}
The sensitivity gap $m-L_F$ governs growth of the error lower bound, while
$KL_F$ bounds growth of the composed score. Holding the other quantities
fixed, a larger gap lowers the distance needed to certify high error, and a
smaller score-growth bound enlarges the guaranteed acceptance radius until
it reaches $r$. The reference residual
can partially cancel the subsequent response mismatch, which accounts for
the subtraction of $\varepsilon_0$ in
Equation~\eqref{eq:app_local_error_bound}. A smaller $\varepsilon_0$ therefore
strengthens this certificate.
The margin $\gamma$ is the available increase in score before rejection,
so a larger margin preserves acceptance over larger deviations.

The result concerns accepted prediction error. A control in $\mathcal A_E$
is a missed OOD detection only if it is outside the conditional training
support. Neither high error nor OOD membership alone proves a physical
safety violation. The argument applies to any threshold with the specified
reference margin. It does not contradict marginal in-distribution coverage
from conformal calibration, which does not guarantee rejection of OOD
queries or accepted predictions with large error.

\subsection{Feasible Directions and Differential Conditions}
\label{app:local_failure_derivatives}

The neighborhood-wide assumptions are sufficient but stronger than needed
to exhibit one failure direction. Fix a unit vector $v$ and suppose
$u_t=u^\ast+tv\in\mathcal U$ for $0\le t\le r$. Here $t$ measures control
displacement, not rollout time. If
\begin{equation}
\|H_O(o,u_t)-H_O(o,u^\ast)\|_2\ge mt,\qquad
\|F_\theta(z,u_t)-F_\theta(z,u^\ast)\|_2\le L_Ft,
\label{eq:app_directional_response}
\end{equation}
the same proof applies with $d=t$. Consequently, every $t$ in the interval
in Equation~\eqref{eq:local_failure_interval} gives a feasible, accepted,
high-error control. This formulation also applies at a control boundary
when the chosen ray remains feasible.

These response bounds can follow from local derivative conditions. If
$H_O(o,\cdot)$ is differentiable at $u^\ast$ and
$\|J_uH_O(o,u^\ast)v\|_2>m$, its first-order expansion gives the true-response
bound in Equation~\eqref{eq:app_directional_response} for a
small $r$. If $F_\theta(z,\cdot)$ is continuously differentiable along the
ray and $\|J_uF_\theta(z,u_t)v\|_2\le L_F$ throughout it, integration gives
\begin{equation}
\|F_\theta(z,u_t)-F_\theta(z,u^\ast)\|_2
\le \int_0^t\|J_uF_\theta(z,u_s)v\|_2\,\mathrm ds
\le L_Ft.
\label{eq:app_derivative_upper_bound}
\end{equation}
A sufficient local condition for the true map is $\inf_{\|v\|_2=1}\|J_uH_O(o,u^\ast)v\|_2>m$, 
for bounds in every direction.
For the learned map, it suffices that
$\|J_uF_\theta(z,u)\|_{\mathrm{op}}\le L_F$ along each segment from $u^\ast$
to a queried control, with the map continuously differentiable on those
segments. 

\subsection{Direct Control Dependence}
\label{app:local_failure_scores}

% \paragraph{Directly control-dependent scores.}
Consider instead $\Psi(z,u)=s_z(u,F_\theta(z,u))$. Suppose that for the
relevant pairs of inputs and predicted successors,
\begin{equation}
|s_z(u,y)-s_z(u',y')|
\le K_u\|u-u'\|_2+K_y\|y-y'\|_2,
\qquad K_u,K_y\ge0.
\label{eq:app_joint_score_lipschitz}
\end{equation}
Then Equation~\eqref{eq:app_local_score_bound} becomes
\begin{equation}
\Psi(z,u)\le\tau_\alpha-\gamma+(K_u+K_yL_F)d.
\end{equation}
The error bound is unchanged, and the proposition holds with acceptance
radius $\min\{r,\gamma/(K_u+K_yL_F)\}$.
% with the same zero-denominator convention. 
Small learned sensitivity alone does not control the direct
term $K_u$. A conditional flow score or ensemble disagreement must satisfy
bounds for its own complete score map.

\section{Experimental Settings and Additional Results}
\label{app:experiments}
\label{appendix:exp}

This section provides the task configurations, model architectures, training schedules, and additional evaluations supporting the main experiments. The evaluations use \(200\) trials for Dubins obstacle avoidance, \(1000\) trials for block plucking, and \(50\) trials for real-robot obstacle avoidance. Throughout, \(\alpha\) denotes nominal in-distribution coverage, so larger \(\alpha\) gives a more permissive acceptance threshold for fixed calibration scores.

\paragraph{Evaluation metrics.}
For obstacle avoidance, \(\rho_{\mathrm{succ}}\) is the fraction of trajectories that remain safe, \(\rho_{\mathrm{filt}}\) is the fraction of reference controls overridden by the filter, and \(\rho_{\mathrm{viol}}\) is the fraction of executed transitions classified as OOD by the evaluated model and surrogate. In manipulation, \(\rho_{\mathrm{fail}}\) measures task failures and \(\rho_{\mathrm{inc}}\) records incomplete executions, distinguishing failure avoidance from task completion.
The next-step latent prediction error \(\mathcal E_z\) in Table~\ref{apptable:dubins_accuracy} compares the predicted prior logits \(F_\theta(z_t,u_t)\) with the posterior logits inferred from the actual next observation, without perturbing \(u_t\).
It is distinct from the two response energies \(\mathcal E^{F_\theta}\) and \(\mathcal E^O\) in Eq.~\eqref{eq:mismatch_metric}.
% , which are plotted separately to compare changes under control perturbations.
OOD violations are defined relative to each fitted surrogate and its calibrated threshold. We assess physical task outcomes separately from surrogate-defined violations and reconstruction quality.

\paragraph{Reconstruction metric.}
\label{app:reconstruction}
Reconstruction uses the observation-conditioned posterior latent, rather than an open-loop prior prediction.
For normalized observed image \(I_{it}^{(c)}\) and its decoded reconstruction \(\hat I_{it}^{(c)}\), we sum squared pixel and channel residuals over the available cameras \(\mathcal C\), then average over timesteps and evaluation rollouts:
\begin{equation}
\mathcal E_{\mathrm{recon}}
=\frac1{N_{\mathrm{roll}}}\sum_{i=1}^{N_{\mathrm{roll}}}
\frac1{T_i}\sum_{t=1}^{T_i}\sum_{c\in\mathcal C}
\|\hat I_{it}^{(c)}-I_{it}^{(c)}\|_F^2.
\end{equation}
Image intensities are scaled by \(1/255\); residuals are summed, not averaged, over image coordinates.

\subsection{Dubins Obstacle Avoidance}
\label{app:dubins}

\paragraph{Environment and data.}
The agent moves at constant speed \(v=1\,\mathrm{m/s}\), with steering-rate control \(u_t\in[-1.25,1.25]\,\mathrm{rad/s}\) and timestep \(\Delta t=0.05\,\mathrm{s}\). The unsafe region is a disk of radius \(0.4\,\mathrm m\) centered at the origin. Observations are bird's-eye-view RGB images of size \(128\times128\times3\). Training data comprise collision-free trajectories collected using expert and random controls. For position \((x_t,y_t)\) and heading \(\theta_t\), the Euler transition is
\begin{equation}
\begin{bmatrix}x_{t+1}\\y_{t+1}\\\theta_{t+1}\end{bmatrix}
=\begin{bmatrix}x_t\\y_t\\\theta_t\end{bmatrix}
+\Delta t\begin{bmatrix}v\cos\theta_t\\v\sin\theta_t\\u_t\end{bmatrix}.
\label{eq:app_dubins_dynamics}
\end{equation}
The unsafe state set is \(\{(x,y,\theta):x^2+y^2\le0.4^2\}\). Writing the physical transition as \(f_{\mathrm{phys}}\), its control Jacobian is
\begin{equation}
J_uf_{\mathrm{phys}}(x_t,y_t,\theta_t,u_t)
=\begin{bmatrix}0&0&\Delta t\end{bmatrix}^{\mathsf T},
\qquad \|J_uf_{\mathrm{phys}}\|_2=\Delta t.
\end{equation}
Thus the one-step heading response does not vanish near the obstacle; under this discretization, control affects position only at a later step. This physical response does not imply a uniform lower bound in image or latent coordinates, since rendering and encoding can attenuate it. The main-text analysis therefore measures the encoded ground-truth response directly.

\begin{figure}[t]
\centering
\includegraphics[width=\linewidth]{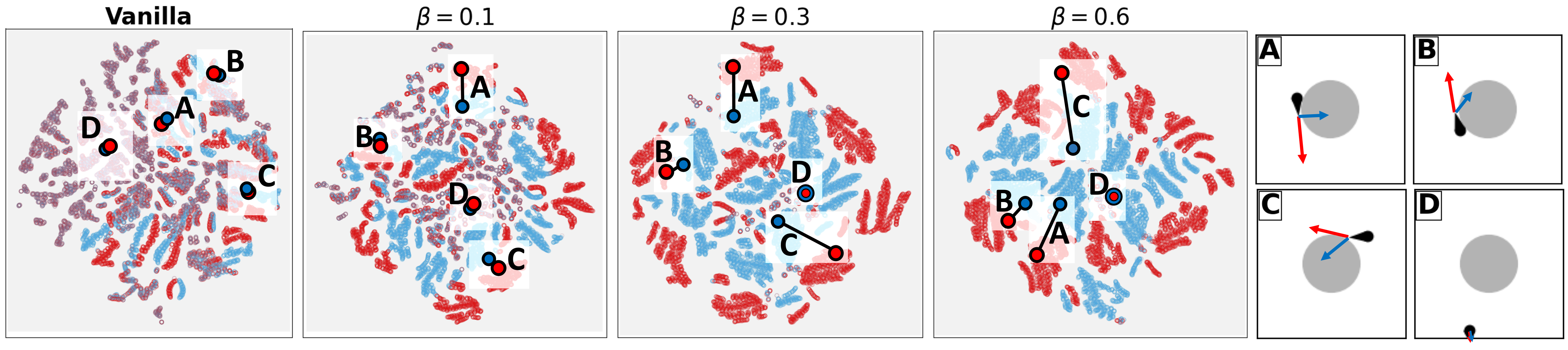}
\caption{\small PCA+t-SNE projections of predicted successor representations under demonstrated controls (red) and undemonstrated controls (blue), for different support fractions \(\beta\). The qualitative separation at \(\beta=0.3\) is already comparable to that at \(\beta=0.6\).}
\label{appfig:dubins}
\end{figure}

\paragraph{Dynamics architecture.}
We use DreamerV3~\cite{hafner2023mastering}. A convolutional encoder with depth \(32\), kernel size \(4\), and minimum spatial resolution \(4\), followed by a five-layer MLP of width \(1024\), maps the image to an embedding \(e_t=\mathrm{Enc}_\theta(o_t)\). Layers use SiLU activations and layer normalization. The RSSM state is \(z_t=(h_t,s_t)\), with deterministic recurrent state \(h_t\in\mathbb R^{512}\) and stochastic state \(s_t\in\mathbb R^{32}\). The image embedding supplies observation information to the posterior \(q_\theta(s_t\mid h_t,e_t)\); it is distinct from the RSSM state. Given \(z_t\) and \(u_t\), the transition computes
\begin{equation}
h_{t+1}=f_\theta(h_t,s_t,u_t),\qquad
s_{t+1}\sim p_\theta(\cdot\mid h_{t+1}).
\end{equation}
Imagined rollouts use this prior, because future observations are unavailable. Training aligns it with the posterior \(q_\theta(s_{t+1}\mid h_{t+1},e_{t+1})\), together with the standard reconstruction and prediction losses. The current conditioning state \(z_t\) uses the full RSSM representation, including the recurrent state and posterior stochastic state; queried successors are produced by the prior in the same coordinates.

\paragraph{Dynamics training and support selection.}
Training runs for \(200{,}000\) gradient steps. Sensitivity regularization begins after \(100{,}000\) steps, following a warm-up that optimizes only \(\mathcal L_{\mathrm{dyn}}\). The high-support subset is refreshed every \(10{,}000\) steps: we recompute latent embeddings, rank training samples by Euclidean \(k\)NN distance with \(k=20\), and retain the fraction \(\beta\) with the smallest scores. Neighbor search uses FAISS on the GPU. The default parameters are \(\beta=0.3\), \(g=8\), and \(\lambda_{\mathrm{reg}}=0.4\), with five Hutchinson probes per training query. Deployment thresholds use nominal coverage \(\alpha=0.90\). Training-time support selection ranks demonstrated samples in the training set.

\paragraph{One-step filtering.}
At each timestep, the filter evaluates the random reference controller's proposed control. It executes that proposal when its score is at most \(\tau_\alpha\); otherwise, it executes the sampled alternative with the smallest score. This replacement need not itself lie below the threshold, so interventions can coexist with nonzero OOD violation rates. The same filtering rule is used with \(k\)NN, flow matching, and ensemble surrogates.

\subsection{Support Selection and Alternative Objectives}
\label{app:ablations}
\label{appendix:rebuttal_exp}

\paragraph{Support-fraction ablation.}
We vary \(\beta\in\{0.1,0.3,0.6\}\), keeping the other Dubins settings fixed. Figure~\ref{appfig:dubins} applies the visualization protocol above. Increasing \(\beta\) produces greater visible separation between demonstrated and undemonstrated controls, with \(\beta=0.3\) already comparable to \(\beta=0.6\).

\begin{table*}[t]
\centering

\begin{minipage}[t]{0.40\textwidth}
\centering
\small
\renewcommand{\arraystretch}{1.12}
\setlength{\tabcolsep}{4pt}
\begin{tabular}{lccc}
\toprule
\textbf{Method} & \(\beta\) & \(\mathcal E_z\downarrow\) 
& \(\mathcal E_{\mathrm{recon}}\downarrow\)\\
\midrule
Vanilla & N/A & \(6.29\times10^{-2}\) & 45.27\\
\midrule
\multirow{3}{*}{Ours}
& 0.1 & \(7.51\times10^{-2}\) & \textbf{31.70}\\
& 0.3 & \(6.14\times10^{-2}\) & 35.88\\
& 0.6 & \(\mathbf{5.92\times10^{-2}}\) & 39.57\\
\bottomrule
\end{tabular}
\captionof{table}{\small Predictive performance for Dubins obstacle avoidance across support fractions.}
\label{apptable:dubins_accuracy}
\end{minipage}
\hfill
\begin{minipage}[t]{0.56\textwidth}
\centering
\small
\renewcommand{\arraystretch}{1.12}
\setlength{\tabcolsep}{4pt}
\begin{tabular}{llcccc}
\toprule
\textbf{Surrogate} & \textbf{Method} & \(\beta\)
& \(\rho_{\mathrm{succ}}\uparrow\)
& \(\rho_{\mathrm{filt}}\)
& \(\rho_{\mathrm{viol}}\downarrow\)\\
\midrule
\multirow{4}{*}{\(k\)NN}
& Vanilla & N/A & 0.600 & 0.340 & 0.273\\
\cmidrule(lr){2-6}
& \multirow{3}{*}{Ours}
& 0.1 & 0.715 & 0.327 & 0.184\\
& & 0.3 & \textbf{0.800} & 0.323 & \textbf{0.166}\\
& & 0.6 & 0.705 & 0.338 & 0.201\\
\midrule
\multirow{4}{*}{FM}
& Vanilla & N/A & 0.910 & 0.309 & 0.153\\
\cmidrule(lr){2-6}
& \multirow{3}{*}{Ours}
& 0.1 & 0.920 & 0.359 & 0.171\\
& & 0.3 & \textbf{0.960} & 0.323 & \textbf{0.104}\\
& & 0.6 & 0.925 & 0.340 & 0.140\\
\bottomrule
\end{tabular}
\captionof{table}{\small Dubins obstacle avoidance with a filtered random reference policy (\(200\) trials).}
\label{apptable:dubins_results}
\end{minipage}

\end{table*}

Table~\ref{apptable:dubins_accuracy} shows a trade-off between next-step latent prediction and reconstruction. Across the regularized models, increasing \(\beta\) lowers \(\mathcal E_z\) from \(7.51\times10^{-2}\) to \(5.92\times10^{-2}\), while reconstruction error increases from \(31.70\) to \(39.57\). Both \(\beta=0.3\) and \(\beta=0.6\) improve next-step latent prediction relative to vanilla; every tested regularized model has lower reconstruction error than vanilla. These results do not establish the cause of the reconstruction trade-off.

Table~\ref{apptable:dubins_results} shows that predictive improvements need not translate monotonically to filtering performance. The default \(\beta=0.3\) gives the highest trajectory success and lowest violation rates among the tested fractions for both \(k\)NN and flow matching. Even \(\beta=0.1\) improves trajectory success over vanilla, although its flow-matching violation rate is higher. Increasing \(\beta\) to \(0.6\) reduces success for both surrogates. These results support selecting a high-support subset rather than assuming that broader regularization is always beneficial.

\paragraph{Action identifiability and forward responsiveness.}
The offline dataset provides outcomes only for executed controls, not counterfactual outcomes for unseen controls at the same states. The proposed regularizer therefore does not rely on ground-truth negative transitions for a contrastive objective. For the inverse-dynamics baselines, we attach an auxiliary head that reconstructs \(u_t\) from the demonstrated latent transition \((z_t,z_{t+1})\), and train it together with the dynamics model. We compare mean-squared-error (MSE) and Gaussian negative-log-likelihood (GNLL) objectives of the form
\begin{align}
\mathcal L_{\mathrm{inv,MSE}}
&=\mathbb E\!\left[\|u_t-\mu_\omega(z_t,z_{t+1})\|_2^2\right],\\
\mathcal L_{\mathrm{inv,GNLL}}
&=\mathbb E\!\left[-\frac1{d_u}\log\mathcal N\!\left(u_t;\mu_\omega(z_t,z_{t+1}),\Sigma_\omega(z_t,z_{t+1})\right)\right],
\end{align}
where expectations are over demonstrated latent transitions. The chosen inverse loss is added to the dynamics-training objective with an auxiliary-loss weight.
The GNLL head follows the same Gaussian architecture and parameterization described in Appendix~\ref{app:ensemble}, with the latent pair as input and controls as the prediction target.
% it predicts \(\mu_\omega\in\mathbb R^{d_u}\) and log standard deviations \(b_\omega\in\mathbb R^{d_u}\), clips \(b_\omega\) to \([-20,1]\), and sets \(\Sigma_\omega=\operatorname{diag}(\exp(2b_\omega))\succ0\).
Its training loss averages the element-wise Gaussian NLL over control coordinates,
% omitting the constant \(\tfrac12\log(2\pi)\);
while the MSE head directly predicts the control vector.
Recovering observed controls can encourage action-relevant representations, but it does not directly impose a lower response on \(F_\theta(z,u)\) as \(u\) varies. Identifiability on demonstrated transitions can therefore coexist with weak forward responses to control changes. The Jacobian penalty directly targets that forward response without requiring counterfactual labels.

\begin{table}[t]
\centering
\small
\renewcommand{\arraystretch}{1.12}
\setlength{\tabcolsep}{4pt}
\begin{tabular}{lccc}
\toprule
\textbf{Method} & \(k\)NN & FM & ENS\\
\midrule
Vanilla & \(0.600/0.273\) & \(0.910/0.153\) & \(0.380/0.327\)\\
Ours (\(\beta=0.3\)) & \(\mathbf{0.800}/\mathbf{0.166}\) & \(\mathbf{0.960}/\mathbf{0.104}\) & \(\mathbf{0.440}/0.297\)\\
Inverse dynamics (GNLL) & \(0.495/0.291\) & \(0.905/0.167\) & \(0.345/0.322\)\\
Inverse dynamics (MSE) & \(0.620/0.283\) & \(0.940/0.136\) & \(0.355/0.328\)\\
Uniform regularization (\(\beta=1\)) & \(0.695/0.207\) & \(0.930/0.129\) & \(0.400/\mathbf{0.290}\)\\
\bottomrule
\end{tabular}
\caption{\small Additional Dubins evaluations with a filtered random reference policy (\(200\) trials). Each entry is \(\rho_{\mathrm{succ}}/\rho_{\mathrm{viol}}\); higher success and lower violation are preferable.}
\label{tab:additional_exp}
\end{table}

Table~\ref{tab:additional_exp} shows that inverse-dynamics benefits vary across surrogates. MSE improves flow-matching success from \(0.910\) to \(0.940\) and lowers violations from \(0.153\) to \(0.136\), but remains below the proposed method's success of \(0.960\). With \(k\)NN, GNLL lowers success from \(0.600\) to \(0.495\), while MSE increases it to \(0.620\). Neither inverse objective matches the proposed method's success across any of the three surrogates. These task-level results do not isolate whether variance prediction, representation geometry, or another training effect explains each difference.

\paragraph{Uniform regularization.}
Setting \(\beta=1\) applies the Jacobian penalty to all training samples. Selective regularization with \(\beta=0.3\) gives higher success for every surrogate and lower violation rates for \(k\)NN and flow matching. Uniform regularization gives a slightly lower ensemble violation rate despite lower success. Thus selective regularization improves the physical task metric in this comparison without dominating every surrogate-defined metric.

\subsection{Training Cost}
\label{app:training_cost}

On a single NVIDIA RTX 5090, \(200{,}000\) Dubins training steps take \(23.1\) hours for the vanilla model and \(27.6\) hours with sensitivity regularization: an increase of \(4.5\) hours, or approximately \(19.5\%\). FAISS-based support-set updates have a separately reported total cost of approximately \(14\) minutes. The regularized run uses the warm-up and refresh schedule in Appendix~\ref{app:dubins}, with regularization active only after step \(100{,}000\). These measurements characterize this configuration; costs depend on dataset size, latent dimension, probe count, and refresh frequency.

\subsection{Block-Plucking Manipulation}
\label{app:manipulation}

\paragraph{Environment.}
In IsaacLab, a Franka arm extracts the middle block from a three-block stack while keeping the top block stably supported by the bottom block. Observations combine wrist-mounted and tabletop RGB cameras, each of size \(256\times256\times3\), with a seven-dimensional proprioceptive input describing the robot and gripper configuration. The control comprises a six-degree-of-freedom end-effector pose increment (three translation and three rotation coordinates) and a discrete gripper open/close command. Continuous control coordinates are normalized to \([-0.5,0.5]\).

\paragraph{Dynamics architecture and training.}
The Dreamer-style RSSM has \(h_t\in\mathbb R^{512}\) and \(32\) categorical stochastic variables, each with \(32\) classes represented by a one-hot vector. The transition updates the recurrent state and categorical logits:
\begin{align}
h_{t+1}&=f_\theta(h_t,s_t,u_t),\qquad
\ell_{t+1}=\ell_\theta(h_{t+1})\in\mathbb R^{32\times32},\\
p_\theta(s_{t+1}\mid h_{t+1})
&=\prod_{j=1}^{32}\mathrm{Cat}\!\left(s_{t+1}^j;\operatorname{softmax}(\ell_{t+1}^j)\right).
\end{align}
Imagined rollouts recursively update \(h_t\) and sample the stochastic state from this prior. Convolutional encoders process images, and a five-layer MLP of width \(1024\) processes proprioception. Networks use SiLU activations and layer normalization. Dynamics training runs for \(250{,}000\) gradient steps, with regularization starting after \(150{,}000\) steps. The support set is refreshed every \(10{,}000\) steps using Euclidean \(k\)NN distances with \(k=20\). The default settings are \(\beta=0.2\), \(g=8\), and \(\lambda_{\mathrm{reg}}=0.2\). Distances for the stochastic component use one-hot representations.
% not categorical logits.

\paragraph{Reachability filtering.}
The ensemble surrogate defines the latent avoid set
\begin{equation}
\mathcal A_\alpha=\left\{z\in\mathcal Z:\min_{u\in\mathcal U}\Psi(z,u)>\tau_\alpha\right\},
\label{eq:app_avoid_set}
\end{equation}
which contains states where every admissible control is classified as OOD. Following~\cite{seo2025uncertainty}, we train a reachability filter from imagined latent rollouts to avoid this set and override reference controls that its learned value predicts will lead toward it. We train a separate reachability policy at each coverage level because changing \(\tau_\alpha\) changes the avoid set. The actor and critic each use two hidden layers of width \(512\) with SiLU activations. The policy is a diagonal Gaussian with learned standard deviations clipped to \([0.1,1.0]\). Training uses Soft Actor-Critic (SAC), imagined-rollout horizon \(15\), up to \(200{,}000\) updates, discount factor \(0.997\), and TD-\(\lambda=0.95\).

\begin{wraptable}{r}{0.45\textwidth}
\vspace{-1.3em}
\centering
\small
\renewcommand{\arraystretch}{1.12}
\setlength{\tabcolsep}{4pt}
\begin{tabular}{lccccc}
\toprule
\textbf{Filter} & \(g\) 
& \(\rho_{\mathrm{fail}}\downarrow\) 
& \(\rho_{\mathrm{inc}}\) 
& \(\rho_{\mathrm{filt}}\) 
& \(\mathcal E_{\mathrm{recon}}\downarrow\)\\
\midrule
Vanilla & N/A & 0.118 & 0.013 & 0.154 & 2207.55\\
\midrule
\multirow{3}{*}{Ours} 
& 5  & 0.073 & 0.029 & 0.360 & \textbf{2134.88}\\
& 8  & \textbf{0.047} & 0.053 & 0.199 & 2146.66\\
& 12 & 0.128 & 0.020 & 0.225 & 2237.44\\
\bottomrule
\end{tabular}
\caption{\small Block-plucking performance across sensitivity targets \(g\).}
\label{apptable:manipulation_margin}
\vspace{-0.8em}
\end{wraptable}

\paragraph{Sensitivity-target ablation.}
\label{app:g_ablation}
Table~\ref{apptable:manipulation_margin} varies the soft target \(g\) for the squared Frobenius norm of the control Jacobian. The default \(g=8\) gives the lowest failure rate among the tested values. Setting \(g=5\) gives the lowest reconstruction error and reduces failure relative to vanilla, but incurs the highest intervention rate. Increasing the target to \(g=12\) worsens both failure and reconstruction. These results show that raising the sensitivity target does not necessarily improve task performance and motivate adaptive selection of \(g\) as future work.

% \Needspace{30\baselineskip}
\subsection{Real-Robot Static-Obstacle Avoidance}
\label{app:hardware}

\paragraph{System and data.}
A Unitree Go2 quadruped executes three discrete commands---turn left, move straight, and turn right---at a fixed forward speed of \(0.5\,\mathrm{m/s}\). The training set consists of \(60\) minutes of collision-free teleoperation. We replace the built-in locomotion interface with a learned low-level RL controller trained using~\cite{unitree2025rllab}, enabling real-time overrides by the safety filter. Two body-mounted cameras provide synchronized first-person RGB images of size \(256\times256\times3\). Images are transmitted over UDP to an offboard workstation with an RTX 5090, which evaluates the learned dynamics and filter and returns the selected command. Camera acquisition and offboard inference run at \(10\,\mathrm{Hz}\).
% giving a control timestep of \(0.1\,\mathrm s\).

\paragraph{Dynamics and filtering.}
The model uses the manipulation architecture's visual encoders and RSSM, with a \(512\)-dimensional recurrent state and \(32\times32\) discrete stochastic state. Training runs for \(200{,}000\) gradient steps, with regularization beginning after \(100{,}000\) steps and settings \(g=6\), \(\lambda_{\mathrm{reg}}=0.2\), \(\beta=0.2\), and \(\alpha=0.95\). Commands are direct one-hot inputs \(u\in\{e_1,e_2,e_3\}\subset\mathbb R^3\), and we evaluate all three admissible commands for control selection. The regularizer differentiates the prior logits with respect to the one-hot input coordinates, holding the current RSSM state fixed.

The \(k\)NN and flow-matching surrogates use one-step filters. The ensemble surrogate uses a reachability filter trained by DQN on imagined latent rollouts~\cite{seo2025uncertainty}. Its Q-network has two hidden layers of width \(100\) and ReLU activations. Training uses \(100{,}000\) updates, discount factor \(0.9999\), and imagination horizon \(15\). The different filter roles matter when comparing ensemble performance across tasks: Dubins tests local candidate rankings directly, whereas manipulation and hardware use values learned over longer imagined trajectories. Observation-level changes can also influence ensemble disagreement on hardware; the experiments do not isolate scene variation as the cause of any particular outcome.